\documentclass[letterpaper]{article}
\usepackage[preprint]{aaai2027}
\usepackage[hyphens]{url}
\usepackage{graphicx}
\usepackage{natbib}
\usepackage{caption}
\usepackage{booktabs}
\usepackage{amsmath}
\usepackage{xcolor}

\title{Beyond Accuracy: Auditing Spatial Provenance in\\Visual Token Pruning for OCR-Critical MLLM Inference}
\author{Feixiang Liu\textsuperscript{1,2}, Qiang Qiu\textsuperscript{1}, Hao Zhang\textsuperscript{1,2}, Xinyue Wang\textsuperscript{1}}
\affiliations{
\textsuperscript{1}State Key Laboratory of AI Safety, Institute of Computing Technology, CAS\\
\textsuperscript{2}University of Chinese Academy of Sciences\\
\texttt{liufeixiang23s@mails.ucas.ac.cn, qiuqiang@ict.ac.cn, zhanghao23s@ict.ac.cn}\\
\texttt{wangxinyue@ict.ac.cn}
}

\begin{document}

\maketitle

\begin{abstract}
Visual-token pruning is usually judged by answer quality at a fixed retention budget. For text-rich multimodal large language models (MLLMs), this protocol can miss a distinct failure: an answer remains correct even when no retained token is locally traceable to the small OCR region that supports it. We turn this blind spot into an evidence-risk audit that couples answer behavior with geometric token-origin provenance, interventions, and realized cost; transparent training-free selectors isolate controlled operating points. On locked image-disjoint confirmation, Qwen Target at 30\% retention has observed accuracy 0.786 versus 0.783 for Full (paired image-cluster difference $+0.003$, 95\% CI [$-0.014,+0.020$]), yet same-budget Target, Random, and Grid retain sharply different positive-support coverage: 0.620, 0.270, and 0.318. Across Qwen3-VL-8B, LLaVA-1.5-7B, and InternVL3.5-8B, matched controls, interventions, detector tests, and external methods reveal model-specific quality--risk--traceability frontiers that accuracy alone does not expose. Materialized prefixes yield up to 4.32$\times$ batch-prefill speedup and 76.4\% lower incremental peak memory; full-validation TextVQA/DocVQA further shows that favorable target-verification points do not imply task-general compression. Visual-token pruning should therefore report surviving spatial provenance and realized cost alongside quality and compression.
Code and reproducibility artifacts are available at \url{https://github.com/SouthWinter/spatial-provenance-audit}.
\end{abstract}

\section{Introduction}

Multimodal large language models (MLLMs) answer questions about images, documents, charts, and scene text through visual-token sequences consumed by a language model \citep{liu2023visualinstructiontuning,bai2023qwenvl,chen2024internvl}. High-resolution text-rich inputs create long prefixes that inflate prefill, motivating visual-token pruning.

FastV, Visual Tokens Withdrawal, FitPrune, G-Prune, VisionZip, and related methods remove, merge, withdraw, or select tokens while preserving the base MLLM \citep{chen2024fastv,lin2025vtw,jiang2025gprune,ye2025fitprune,yang2026visionzip}. They chiefly ask how much task quality remains at a given budget. This is necessary but incomplete: random retention, duplication, unstable scores, and benchmark artifacts can make a short prefix appear reliable after locally attributable support is lost \citep{wen2025rightproblem,wen2025dart}.

OCR-critical prompts expose the gap. A receipt, screenshot, or form query may depend on one tiny word box. After its token cells are removed, the MLLM may still answer from language priors, layout, nearby text, or mixed context. Accuracy can therefore remain correct although no retained cell represents the spatial source.

OCR annotations make this hidden failure measurable. TextOCR associates strings with boxes \citep{singh2021textocr}, allowing the same pruned prefix to be tested for answer behavior and geometric traceability to its supporting region. We call this second quantity \emph{spatial provenance}: the source cells represented by retained or merged visual tokens. It is deliberately narrower than causal evidence, but it answers a question that accuracy cannot---whether the compressed prefix still exposes an inspectable spatial path to the annotated support.

We make the question executable through an evidence-risk protocol and a shared budgeted-selection contract. Transparent training-free selectors vary target relevance, grid coverage, and optional OCR/layout priors while leaving text tokens, model weights, prompts, and decoding unchanged. They expose controlled points on the quality--risk--traceability frontier, whose masks are evaluated jointly by answer behavior, semantically separated region coverage, interventional diagnostics, and measured execution.

The audit reveals a gap hidden by conventional reporting. On locked Qwen confirmation, Target retains 30\% of visual tokens and has observed accuracy within 0.3 points of Full, while same-budget Random and Grid retain only 0.270/0.318 positive-region coverage versus 0.620 for Target. LLaVA and InternVL occupy different frontiers, and open QA requires substantially more retention, showing why a successful operating point cannot be transferred by keep ratio alone. These findings elevate spatial traceability from an unreported implementation detail to an explicit evaluation dimension.

We make three contributions:
\begin{itemize}
\item We identify localized spatial-provenance loss as a failure mode hidden by accuracy--compression evaluation and formulate an evidence-risk protocol that jointly audits answer behavior, spatial traceability, interventions, and realized efficiency.
\item We build a controlled testbed comprising TextOCR-Hard development and locked image-disjoint confirmation sets, semantically separated provenance metrics, and a common budget contract with transparent reference selectors for target relevance, grid coverage, and optional OCR/layout support.
\item We establish across three MLLMs that answer quality and token count do not determine surviving spatial provenance, and characterize the resulting model- and task-specific frontiers through matched controls, external methods, construct-validity tests, detector and position policies, open QA, and measured GPU execution.
\end{itemize}

\section{Related Work}

\noindent\textbf{Training-free visual token removal.}
Visual-token pruning extends dynamic vision-transformer sparsification and merging \citep{rao2021dynamicvit,bolya2022tome}. MLLM methods prune by early attention, deeper withdrawal, recipe search, or graph importance \citep{chen2024fastv,lin2025vtw,ye2025fitprune,jiang2025gprune}; QuietPrune, HAWK, IF-Prune, and RTPrune add query, head, information-flow, or OCR-specific guidance \citep{gao2026quietprune,zhu2026hawk,sun2026ifprune,wan2026rtprune}. They shorten visual prefixes substantially, but usually emphasize aggregate quality or throughput rather than survival of a task-critical region.

\noindent\textbf{Compression beyond naive top-$k$.}
A second line preserves more structure than independent score-and-drop selection. Merge-based methods aggregate dominant and contextual tokens, while projector-level compression jointly controls redundancy and spatial sparsity \citep{shang2026llavaprumerge,yang2026visionzip,li2024tokenpacker,wu2026vlmpruner}. Structure-aware methods preserve attention outputs or exploit document layout and query relevance \citep{tan2025tokencarve,choi2026docprune}. Coverage-oriented selectors then combine relevance with feature coverage, diversity, complementary context, or protected anchors \citep{zhang2025cdpruner,deng2025scope,xu2026score,li2025mob,tan2026idpruner,wang2026posprune,du2026coin,wang2026tops,oh2026anchorprune}. These mechanisms optimize compressed-task quality, but do not by themselves determine whether a localized source remains spatially traceable. We audit that distinction through positive support, negative-source retention, interventions, and realized speed, with matched CoIn, SCOPE, FastV, VisionZip, and AnchorPrune comparisons. Deletion methods restore native order; VisionZip retains contextual merging.

\noindent\textbf{Text-rich grounding and pruning reliability.}
TextVQA, DocVQA, TextOCR, and OCRBench show why local evidence matters: an answer may depend on one small word, field, or label \citep{singh2019textvqa,mathew2020docvqa,singh2021textocr,liu2024ocrbench}. PinPoint and spatial studies show that compression can preserve global semantics while damaging local integrity \citep{kwon2026pinpoint,guo2025crop,kamath2023whatsup,rajabi2024gsrbench}. N\"uwa addresses grounding collapse by retaining global spatial anchors and then pruning text-guided tokens inside the language model \citep{huang2026nuwa}; GAP, Reroute, and DSTP likewise expose position or irreversible-deletion effects \citep{chien2025gap,yang2026reroute,kim2026dstp}. N\"uwa is the closest method-level overlap: it protects a global reference frame for grounding quality, whereas we audit whether annotation-defined positive support and confusable-negative source regions remain spatially attributable, behaviorally adequate, and efficient after pruning.

UniPruneBench finds random pruning strong and OCR vulnerable; related analyses expose duplication, unstable importance, benchmark bias, and deeper OCR information horizons \citep{peng2025uniprunebench,wen2025rightproblem,wen2025dart,wang2026informationhorizon}. Calibration interventions further link kept-set coverage to confidence \citep{tan2026calibration}. Our contribution unifies positive/negative provenance semantics, locked hard negatives, interventions, position and detector controls, measured cost, and open-QA scope in one pruning audit.

\section{Problem Setup}

An OCR-critical pruning instance consists of an image $I$, a text prompt $p$ that asks whether a target string is present, and a yes/no answer space. A positive probe queries the annotated source word; its box is answer-supporting evidence. A near-miss negative queries a confusable but absent string; the annotated box then localizes only the source word from which that decoy was constructed. Retaining this source region may provide local counter-evidence, but cannot prove that the queried string is absent everywhere in the image. The frozen MLLM converts the image into projected visual embeddings $Z=\{z_i\}_{i=1}^{N}$ and the prompt into text-token embeddings $Q=\{q_j\}_{j=1}^{M}$. Our deletion selectors use keep ratio $\rho$, set $K=\lceil\rho N\rceil$, and return
\begin{equation}
\label{eq:pruning-policy}
S=\pi(Z,Q;\rho),\qquad |S|=K .
\end{equation}
They produce the ordered output sequence $O=(z_i)_{i\in S}$ in original image-token order. More generally, a compression operator produces $K$ output tokens $O=(o_u)_{u=1}^{K}$. Each output has a source-lineage set $L_u\subseteq\{1,\ldots,N\}$ and a representative source anchor $a_u\in L_u$. Deletion gives $L_u=\{a_u\}=\{i\}$ for a retained index $i$; a merged output may have $|L_u|>1$. The model receives $O$ followed by all text tokens and produces the yes/no decision without changing the MLLM.

The pruning module changes neither weights, prompts, labels, nor decoding. Box-free selectors use only the image and prompt; box-aware variants are assisted when supplied by an application-side OCR/layout detector and diagnostic when supplied with annotations. This distinguishes deployable box-free and detector-assisted selection from oracle diagnostics.

For annotated regions $R=\{R_m\}$, each original token has a grid-induced provenance cell $B_i$. It records token origin, not a strict receptive-field boundary: tokens outside $R$ may still encode its information \citep{fan2026visualtokens}. Let $\mathcal{L}(O)=\cup_{u=1}^{K}L_u$ be the source lineage represented by the compressed sequence and $\mathcal{A}(O)=\{a_u\}_{u=1}^{K}$ its representative anchors. We define lineage-based geometric coverage
\begin{equation}
\label{eq:ecr}
\mathrm{C}_{\mathrm{L}}(O,R)=
\frac{\mathrm{area}\left((\cup_m R_m)\cap(\cup_{i\in\mathcal{L}(O)} B_i)\right)}
{\mathrm{area}\left(\cup_m R_m\right)}.
\end{equation}
Anchor coverage $\mathrm{C}_{\mathrm{A}}(O,R)$ replaces $\mathcal{L}(O)$ by $\mathcal{A}(O)$ in Eq.~(\ref{eq:ecr}).
Let $\mathcal{D}^{+}$ and $\mathcal{D}^{-}$ denote positive and near-miss-negative probes, with regions $R_x^{+}$ and $R_x^{-}$. We report two semantically distinct averages:
\begin{equation}
\label{eq:semantic-coverage}
\begin{aligned}
\mathrm{PosECR}&=\frac{1}{|\mathcal{D}^{+}|}\sum_{x\in\mathcal{D}^{+}}\mathrm{C}_{\mathrm{L}}(O_x,R_x^{+}),\\
\mathrm{NegSRC}&=\frac{1}{|\mathcal{D}^{-}|}\sum_{x\in\mathcal{D}^{-}}\mathrm{C}_{\mathrm{L}}(O_x,R_x^{-}).
\end{aligned}
\end{equation}
PosECR measures geometric source-lineage availability for answer-supporting regions. NegSRC measures source-lineage retention of the confusable source region, not evidence of global absence. For merging baselines we additionally average $\mathrm{C}_{\mathrm{A}}$ as AnchorECR, which uses only representative output locations and therefore measures local traceability rather than source participation. Neither quantity is an accuracy surrogate, proof of causal use, or guarantee that merged features preserve separable information from every source. Detailed diagnostics sometimes report the PosECR/NegSRC macro-average as descriptive mean source coverage; we neither optimize it as a unified objective nor interpret high NegSRC as intrinsically beneficial. CenterR and PatchR are supplementary.

\section{Audit Protocol and Reference Selectors}

\begin{figure*}[t]
\centering
\includegraphics[width=\textwidth]{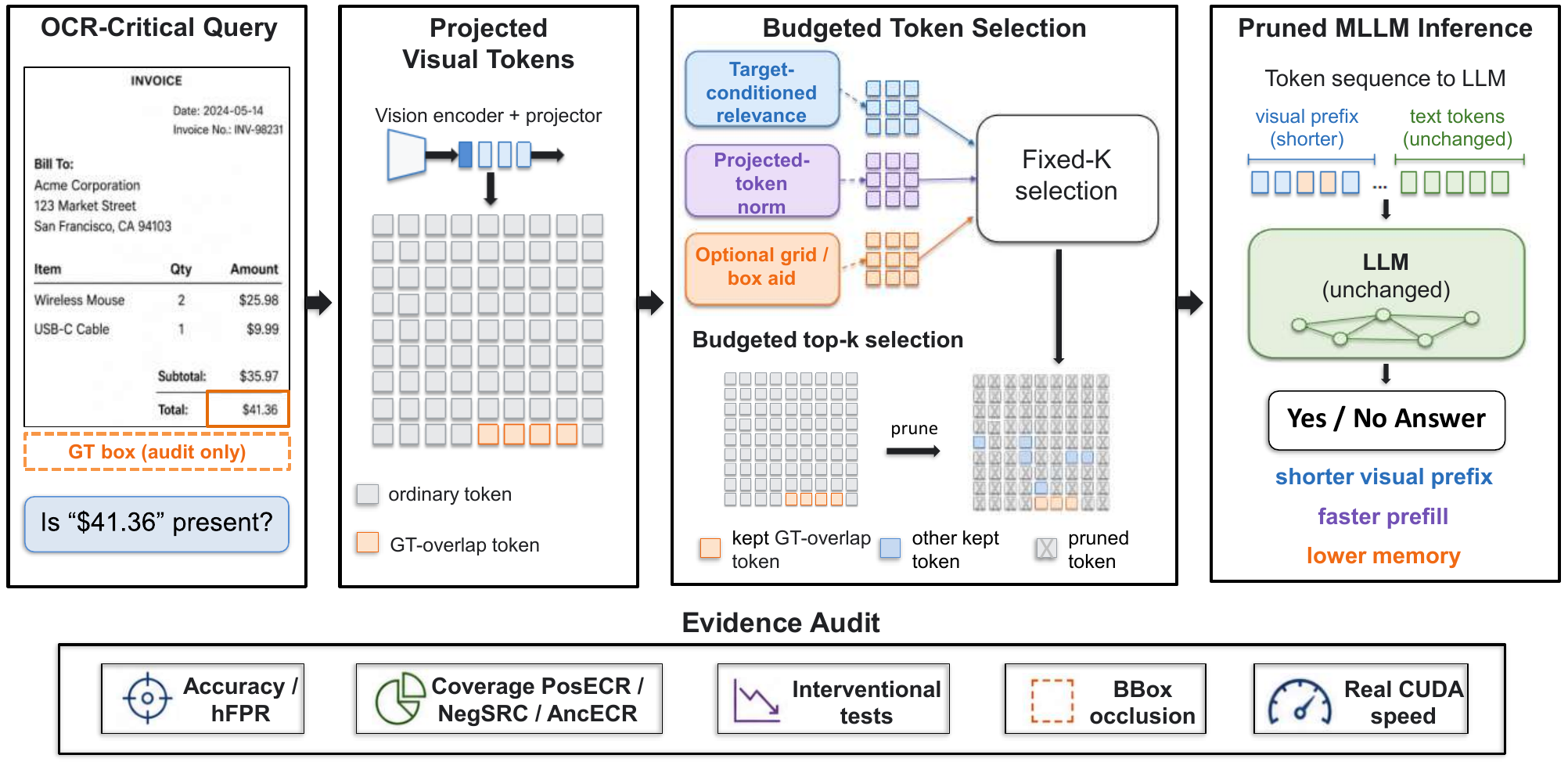}
\caption{Evidence-risk audit overview. Box-free Target selects a fixed-budget prefix using target-conditioned relevance and projected-token norm; assisted variants optionally add grid reservation or supplied OCR/layout-box priors or reservations. The orange GT box and overlap labels are post hoc audit annotations, not Target inputs. We evaluate the shortened prefix for answer behavior, spatial provenance, interventions, and realized CUDA cost.}
\label{fig:overview}
\end{figure*}

\subsection{Overview and Design Requirements}

Figure~\ref{fig:overview} follows the query through projected tokens, budgeted selection, and shortened-prefix inference. The framework has two roles: controlled selectors produce comparable masks from visual and prompt embeddings, a budget, and optional OCR/layout boxes; the audit then evaluates each ordered subset against answer behavior, annotated regions, hard negatives, interventions, and runtime. This separation makes pruning mechanisms comparable under a shared reporting contract.

\subsection{Budgeted Selection Contract}

Let $K=\lceil\rho N\rceil$. For a score vector $s$, candidate set $A$, and count $k$, let $\mathrm{Top}(s,k;A)$ return the $k$ highest-scoring indices in $A$, with token index breaking ties. Every variant first constructs a reserved set $P$ and then applies the same completion rule
\begin{equation}
\label{eq:selection-contract}
S=P\;\cup\;\mathrm{Top}(s,K-|P|;\{1,\ldots,N\}\setminus P).
\end{equation}
Variants differ only in score $s$ and reservation $P$. Target uses $s=a$ and $P=\emptyset$. Target+Grid reserves $P=\mathrm{Grid}(B,\lfloor0.5K\rceil)$, which selects patches nearest evenly spaced image targets; pure Grid sets $S=\mathrm{Grid}(B,K)$. Protected reserves at most $\lfloor0.5K\rceil$ evidence patches, ordered per box by normalized overlap then relevance and selected round-robin. Center-protected reserves each box's nearest center-inside patch, or the nearest patch as fallback. Soft evidence changes only $s$ below. Without boxes, box-aware variants reduce to Target.

This deterministic contract isolates three controllable factors before generation---target relevance, coarse spatial coverage, and evidence availability---while keeping the budget and materialization path fixed. Answer risk is then measured on the resulting prefix without adding a model call to selection.

\subsection{Target-Conditioned Token Scoring}

The box-free selector combines established projected-token norm and visual--text alignment signals \citep{du2026coin}, but conditions their aggregation on the queried target span $Q^\star\subseteq Q$. Original-question experiments use all question tokens or answer-free \emph{Focus}, at most eight unique content words after removing fixed interrogatives and boilerplate; neither sees answers, OCR text, or boxes. Let $c_i^{(\ell)}$ be the $\ell$-th largest similarity in $\{\cos(z_i,q):q\in Q^\star\}$. Then
\begin{equation}
\label{eq:target-relevance}
r_i=\frac{1}{m}\sum_{\ell=1}^{m}c_i^{(\ell)},\quad m=\min(2,|Q^\star|),
\end{equation}
so a token can align with the strongest one or two query tokens rather than with the entire prompt. The selection score is
\begin{equation}
\label{eq:selection-score}
a_i=0.85\,\mathrm{minmax}(r_i)+0.15\,\mathrm{minmax}(\|z_i\|_2),
\end{equation}
where $\mathrm{minmax}(x_i)=(x_i-\min_jx_j)/(\max_jx_j-\min_jx_j)$ per image when the range is nonzero, and $0$ otherwise. Top-two aggregation avoids averaging over the full prompt; the fixed $0.85/0.15$ mixture keeps relevance dominant and uses token norm as a weak projector-space prior. Constants are not retuned per example.

\subsection{Evidence-Aware Token Selection}

Target top-$k$ applies Eq.~(\ref{eq:selection-contract}) with no reservation and uses only the image and prompt. Target+Grid uses the fixed half-budget reservation above, then fills remaining slots by target relevance to hedge against concentrated scores.

With annotation or detector boxes, Protected reserves overlap-balanced evidence tokens before relevance filling; Center-protected anchors each box center as a stricter diagnostic. Their explicit reservation provides a controlled high-coverage endpoint against which soft priors and box-free selection can be compared.

Soft evidence instead scores
\begin{equation}
\label{eq:soft-evidence-score}
b_i=\mathrm{minmax}(a_i)+\beta h_i,
\end{equation}
where $h_i=\max_m\{\mathrm{area}(B_i\cap R_m)/\mathrm{area}(B_i),\,\mathbf{1}[\mathrm{center}(B_i)\in R_m]\}$. We fix $\beta=0.05$, a small evidence boost rather than hard reservation.
The supplement tabulates each variant's intended deployment role and measured operating characteristics.

\subsection{Inference Workflow and Complexity}

The routine computes scores, reserves optional grid/evidence tokens, fills the budget, and restores image-token order. It costs an $N\times|Q^\star|$ similarity matrix and top-$k$, plus spatial lookup for assisted variants; timing includes scoring and materialization.

\subsection{Materialization and Evidence Audit}

Materialization changes token presence but not weights, retained-token order, prompt, or decoding. LLaVA and InternVL main results compact logical position IDs as in their ordinary physically shortened forward paths; Qwen retains the model-derived multimodal RoPE IDs of selected tokens. Physical cache slots remain contiguous. Because position handling can itself affect grounding \citep{chien2025gap}, we hold masks fixed and repeat the LLaVA and InternVL main policies with pre-pruning logical IDs in the supplement.

\paragraph{Evidence-risk reporting contract.}
Each result identifies model, selector, budget, position policy, and detector source, then reports answer behavior, geometric provenance, interventions, and realized cost. This separates availability from causal use and nominal token count from measured efficiency.

\section{Experiments}

\subsection{Experimental Setup}

\paragraph{Datasets.}
TextOCR-Hard is built from TextOCR validation images \citep{singh2021textocr}. From each image we select one uniquely occurring OCR word of length 4--18 whose normalized box area is $5{\times}10^{-5}$--$1.2{\times}10^{-3}$, sampling among the eight smallest eligible words. Its exact string forms a positive probe; a confusable substitution or deletion forms a negative probe and is rejected if the decoy occurs in the same-image OCR. Development and locked confirmation sets each contain 500 images and 1000 paired probes, with no image overlap; seeds, hashes, and operating points were frozen before confirmation inference. OCRBench, TextVQA, DocVQA, and GSR-Bench provide transfer boundaries \citep{liu2024ocrbench,singh2019textvqa,mathew2020docvqa,rajabi2024gsrbench}. Original-question generation uses all 5000 TextVQA and 5349 DocVQA validation samples; matched 500-sample splits support ablations.

\paragraph{Models.}
We evaluate Qwen3-VL-8B-Instruct, LLaVA-1.5-7B-HF, and InternVL3.5-8B-HF \citep{bai2025qwen3vl,liu2023visualinstructiontuning,wang2025internvl35}. Qwen and LLaVA selectors and budgets were chosen on the 1000 development probes, then frozen for the image-disjoint confirmation set. InternVL uses an image-group-disjoint hash split that keeps paired probes from each image together: the Full-derived development threshold $t=2.043$ is applied unchanged to every main-table row, and all results are evaluated on 536 held-out probes. Selector-specific accuracy-optimal and risk-constrained thresholds are supplementary calibration diagnostics; none changes masks or provenance.

\paragraph{Baselines.}
We compare full inference, random, grid, and shuffled-score controls. For LLaVA, adaptations of the official FastV and VisionZip implementations span 20--50\% budgets \citep{chen2024fastv,yang2026visionzip}.
Qwen uses a native VisionZip adaptation at 30\%. SCOPE uses its public LLaVA implementation \citep{deng2025scope}; AnchorPrune uses a parity-checked adaptation of its public implementation, and CoIn is reimplemented from its published algorithm with the reported LLaVA setting \citep{oh2026anchorprune,du2026coin}. These external methods are compared at 40\%. The supplement documents implementation parity, merge-aware lineage, scope boundaries, and budget curves; cross-backbone comparisons are restricted to native or explicitly identified mechanism-level implementations.

\paragraph{Metrics.}
We report accuracy, hard-negative false-positive rate (hFPR) \citep{li2023pope}, threshold-free AUROC, keep ratio, PosECR, and NegSRC. Localization precision (LPP) and its harmonic mean with ECR (Geo-F1) diagnose coarse-cell inflation. Open generation uses exact match, TextVQA soft accuracy, or DocVQA ANLS. Original-question selectors receive only the question, never the answer. Paired intervals, detector robustness, interventions, bounding-box occlusion, and GPU timing complete the audit.

\paragraph{Image-dependence controls.}
We replace each matched image by a same-sized gray image or by an unrelated TextOCR image under three deterministic derangements. Each positive/negative pair receives the same wrong image; fixed points and target/source-string collisions under seven normalizers are excluded. Qwen and LLaVA use all 1000 locked confirmation probes. InternVL uses the fixed 536-probe held-out split and its Full-derived development threshold, without recalibration. We also stratify cases whose decoy occurs as a real OCR token in another TextOCR image.

\paragraph{Implementation.}
Experiments use one NVIDIA A800 80GB GPU, PyTorch 2.6.0 with CUDA 12.4, and Transformers 4.57.1. Each binary prompt is ``Does the image contain the exact text \emph{target}? Answer yes or no.'' Under the model's chat template, we score the complete continuations `` yes'' and `` no'' and predict yes when $\ell_{\mathrm{no}}-\ell_{\mathrm{yes}}\geq t$; $t=0$ for Qwen/LLaVA, while InternVL selects $t$ on its development split as specified above. Open answers use greedy generation with at most 32 new tokens. Qwen inputs are constrained to 802,816 pixels and produce 731--840 visual tokens on confirmation; LLaVA uses $336\times336$ inputs and 576 tokens; InternVL dynamic tiling produces 768--3328 tokens (mean 2494). Random controls use recorded deterministic seeds, and timing reports warm-up, repetitions, batch size, and whether OCR latency is included.

\paragraph{Primary estimands and analysis hierarchy.}
The primary behavioral estimand is the paired confirmation-set accuracy difference between Qwen Target (30\%) and Full; the primary geometric estimands compare its PosECR with matched-budget Random and Grid. Confidence intervals use 10,000 paired bootstrap draws after averaging probe-level differences within each of the 500 images; open-QA intervals resample examples. The locked Random row conditions on one prespecified mask seed; five additional seeds, fixed jointly without outcome selection, form a post-lock confirmation sensitivity analysis. We read behavior and provenance jointly, because neither alone establishes reliable pruning. Other models, thresholds, budgets, interventions, tasks, and metrics are diagnostic or exploratory rather than a multiplicity-controlled confirmatory family.

\begin{table*}[t]
\centering
\small
\begin{tabular*}{\textwidth}{@{\extracolsep{\fill}}lllrrrrrrr@{}}
\toprule
Model & Method & Type & Keep & Acc. & hFPR & AUROC & PosECR & AncECR & NegSRC \\
\midrule
\multicolumn{10}{l}{\textit{A. Box-free methods and controls}} \\
Qwen3-VL-8B & Full & Full & 1.000 & 0.783 & 0.248 & 0.856 & 1.000 & 1.000 & 1.000 \\
Qwen3-VL-8B & Target (30\%) & Box-free ref. & 0.301 & 0.786 & 0.240 & 0.855 & 0.620 & 0.620 & 0.508 \\
Qwen3-VL-8B & Random (30\%) & Box-free ctrl. & 0.301 & 0.739 & 0.160 & 0.776 & 0.270 & 0.270 & 0.300 \\
Qwen3-VL-8B & Grid (30\%) & Box-free ctrl. & 0.301 & 0.723 & 0.142 & 0.755 & 0.318 & 0.318 & 0.318 \\
Qwen3-VL-8B & VisionZip (30\%) & Box-free ext. & 0.301 & 0.748 & 0.198 & 0.788 & 1.000 & 0.846 & 1.000 \\
LLaVA-1.5-7B & Full & Full & 1.000 & 0.625 & 0.316 & 0.666 & 1.000 & 1.000 & 1.000 \\
LLaVA-1.5-7B & Random (40\%) & Box-free ctrl. & 0.401 & 0.669 & 0.256 & 0.712 & 0.372 & 0.372 & 0.389 \\
LLaVA-1.5-7B & Target (40\%) & Box-free ref. & 0.401 & 0.624 & 0.574 & 0.713 & 0.471 & 0.471 & 0.443 \\
LLaVA-1.5-7B & SCOPE (40\%) & Box-free ext. & 0.401 & 0.586 & 0.562 & 0.644 & 0.614 & 0.614 & 0.614 \\
LLaVA-1.5-7B & AnchorPrune (40\%) & Box-free ext. & 0.401 & 0.578 & 0.586 & 0.642 & 0.635 & 0.635 & 0.639 \\
LLaVA-1.5-7B & CoIn (40\%) & Box-free ext. & 0.401 & 0.579 & 0.674 & 0.664 & 0.666 & 0.666 & 0.680 \\
InternVL3.5-8B & Full & Full & 1.000 & 0.646 & 0.306 & 0.687 & 1.000 & 1.000 & 1.000 \\
InternVL3.5-8B & Random (50\%) & Box-free ctrl. & 0.500 & 0.604 & 0.485 & 0.677 & 0.743 & 0.743 & 0.740 \\
InternVL3.5-8B & Grid (50\%) & Box-free ctrl. & 0.500 & 0.623 & 0.474 & 0.699 & 0.695 & 0.695 & 0.695 \\
\midrule
\multicolumn{10}{l}{\textit{B. Annotation-box oracle diagnostics (not deployable rankings)}} \\
LLaVA-1.5-7B & Protected (40\%) & Oracle GT & 0.401 & 0.643 & 0.352 & 0.694 & 1.000 & 1.000 & 1.000 \\
InternVL3.5-8B & Soft evidence (50\%) & Oracle GT & 0.500 & 0.616 & 0.530 & 0.696 & 0.902 & 0.902 & 0.883 \\
\bottomrule
\end{tabular*}
\caption{Primary TextOCR-Hard results; AUROC is threshold-free. Qwen/LLaVA use locked confirmation ($n=1000$, $t=0$); InternVL uses its held-out test ($n=536$, common Full-derived $t=2.043$). Panel A is box-free; Panel B isolates annotation-box oracle diagnostics; detector-assisted results are separate. PosECR uses source lineage, AncECR uses representative output anchors, and NegSRC is confusable-source coverage rather than target-absence evidence. Rows are deployment-specific frontier points, not a ranking.}
\label{tab:main}
\end{table*}

\begin{table*}[!t]
\centering
\small
\begin{tabular*}{\textwidth}{@{\extracolsep{\fill}}llrrrr@{}}
\toprule
Model & Image input & $n$ & Acc. & hFPR & AUROC \\
\midrule
Qwen3-VL-8B & Matched / Blank / Mismatch & 1000 & 0.783 / 0.502 / 0.503 & 0.248 / 0.002 / 0.012 & 0.856 / 0.574 / 0.399 \\
LLaVA-1.5-7B & Matched / Blank / Mismatch & 1000 & 0.625 / 0.503 / 0.556 & 0.316 / 0.002 / 0.234 & 0.666 / 0.693 / 0.568 \\
InternVL3.5-8B & Matched / Blank / Mismatch & 536 & 0.646 / 0.618 / 0.525 & 0.306 / 0.582 / 0.102 & 0.687 / 0.676 / 0.524 \\
\midrule
\multicolumn{6}{l}{\textit{Positive-region metric validity on matched images}} \\
\multicolumn{2}{l}{Policy and diagnostic split} & $n$ & ECR & LPP & Geo-F1 \\
\midrule
\multicolumn{2}{l}{Qwen Target (30\%), development positives} & 500 & 0.651 & 0.205 & 0.295 \\
\multicolumn{2}{l}{LLaVA Target (40\%), development positives} & 500 & 0.458 & 0.112 & 0.171 \\
\multicolumn{2}{l}{InternVL Soft evidence (50\%), development positives} & 500 & 0.916 & 0.166 & 0.255 \\
\midrule
\multicolumn{6}{l}{\textit{Measured GPU efficiency on materialized shortened prefixes}} \\
\multicolumn{2}{l}{Setting} & Keep & Single & Prefill & Memory \\
\midrule
\multicolumn{2}{l}{Qwen Target (20\%)} & 0.200 & 1.45$\times$ & 4.32$\times$ & $-$76.4\% \\
\multicolumn{2}{l}{Qwen Target (30\%)} & 0.301 & 1.37$\times$ & 3.12$\times$ & $-$66.8\% \\
\multicolumn{2}{l}{LLaVA Protected (40\%)} & 0.401 & 1.24$\times$ & 2.34$\times$ & $-$56.4\% \\
\multicolumn{2}{l}{InternVL Target / Soft (50\%)} & 0.500 & 1.57 / 1.54$\times$ & 2.25 / 2.24$\times$ & $-$49.8\% \\
\bottomrule
\end{tabular*}
\caption{Core validity and efficiency evidence. Image entries use Full prefixes and report Matched/Blank/three-seed Mismatch means under fixed matched-image thresholds. Geometry rows are development-set construct diagnostics for the named policies, not the locked-confirmation PosECR values; LPP penalizes retained area outside the region. Efficiency reports measured single-sample and batch-prefill speedup and incremental peak-memory change; batch sizes are 32/100/50 for Qwen/LLaVA/InternVL.}
\label{tab:validity}
\end{table*}

\subsection{Answer Quality Does Not Determine Spatial Provenance}

Table~\ref{tab:main} exposes the evaluation gap directly. At the same 30\% Qwen budget, Target gains +0.047/+0.063 accuracy and +0.350 ([+0.304,+0.394]) / +0.302 ([+0.259,+0.346]) PosECR over Random/Grid, but its hFPR is also higher (0.240 vs.\ 0.160/0.142). Thus token count fixes neither answer behavior nor surviving spatial provenance, and Target occupies a different quality--risk--traceability frontier point rather than uniformly dominating the controls.

Relative to Full, Target's observed accuracy is 0.786 versus 0.783, a paired difference of +0.003 (image-cluster 95\% CI [-0.014,+0.020]) while removing 70\% of visual tokens. Because no noninferiority margin was prespecified, the interval bounds the observed change rather than proving equivalence. Native VisionZip provides a complementary crossing: at the same budget it reaches 0.748 accuracy and 1.000 lineage PosECR, while Target gains +0.038 accuracy ([+0.018,+0.059]) but loses 0.380 lineage coverage ([-0.414,-0.348]). VisionZip's AnchorECR is 0.846, distinguishing exhaustive source participation in a merge from locally traceable output positions.

The frontier is backbone-dependent. On LLaVA confirmation, annotation-box-oracle Protected retains every annotated region, whereas Random favors answer metrics at 0.372 PosECR; box-free SCOPE raises PosECR over Target (0.614 vs. 0.471) with lower accuracy. Under InternVL's common Full-derived threshold, annotation-box-oracle Soft evidence raises PosECR from Grid's 0.695 to 0.902 but trails it by 0.007 accuracy and raises hFPR from 0.474 to 0.530; their AUROCs are 0.696 and 0.699. Calibration is also backbone-specific: Qwen Target/Full AUROC is 0.855/0.856; LLaVA Target reaches 0.713/0.666 but retains 0.442 hFPR under the Full-development threshold. Thus availability, ranking, and risk can move separately.

External selectors occupy different frontier points. On LLaVA confirmation, CoIn differs from SCOPE by only $-0.007$ accuracy but adds 0.112 hFPR ($p<10^{-4}$). Target gains 0.038 accuracy over SCOPE, with unresolved hFPR difference (+0.012, CI [$-$0.034,0.056]); SCOPE's pure-coverage ablation adds 0.086 hFPR ([0.052,0.120]) without an accuracy or PosECR gain.

\subsection{Does Spatial Provenance Capture a Real Failure?}

Conditioning on correctly answered positives isolates the hidden failure from ordinary model errors. On confirmation, Qwen Target reduces low/zero PosECR to 27.8\%/13.1\%, versus 71.2\%/37.0\% for Random. LLaVA Protected eliminates geometric loss; AnchorPrune has 33.2\%/18.1\% low/zero PosECR.

Across four Qwen masks at the same 30\% confirmation budget, ECR is associated with selected-minus-Full yes-margin after controlling box area, cell area, Full margin, and method ($r_s=0.232$ [0.191,0.273]); image and method fixed effects give 0.237 [0.186,0.286]. Complementary development interventions show that removing annotated-region tokens lowers accuracy from 0.798 to 0.769; restoring 25\% recovers 62.1\% of the loss, versus at most 10.3\% for matched random restoration. ECR correlates with adjusted occlusion and deletion drops ($r_s=0.334/0.308$); after controlling box area, cell area, and Full margin, deletion remains associated at 0.288 [0.203,0.371], whereas occlusion becomes 0.102 [$-$0.082,0.285].

On 102 human-validated rendered single-character replacements, substituted-glyph boxes occupy 19.1\% of the word box on average. Target reaches 0.912 positive-query accuracy and 0.897 edit-region coverage, versus 0.716/0.311 for Random and 0.588/0.411 for Grid; its paired coverage gains are +0.586 [0.492,0.676] and +0.486 [0.397,0.575]. This all-render audit tests mask localization; a stricter original/replacement/sham/erase behavioral test passes in 26/102 Qwen cases.

Wrong-image accuracy falls from 0.783 to 0.503 for Qwen, 0.625 to 0.556 for LLaVA, and 0.646 to 0.525 for InternVL (Table~\ref{tab:validity}); every per-seed interval excludes zero. Yet ECR remains recall-like: on development positives, InternVL's 0.916 ECR corresponds to 0.166 LPP and 0.255 Geo-F1. We therefore compare within backbone, policy, budget, and split. Conditional analysis supports convergent validity for Qwen deletion, not a backbone-general causal surrogate; LLaVA's and InternVL's intervals include zero. On 32 double-labeled open-QA cases, union IoU is 0.389 but answer-value presence F1 is 0.968, leaving fine context boundaries uncertain.

Across the prespecified Random row and five fixed post-lock replicates, accuracy is $0.733{\pm}0.005$ (0.726--0.739) and PosECR $0.294{\pm}0.016$ (0.270--0.315); a seed-and-image bootstrap gives Target--Random +0.053 accuracy [0.033,0.073] and +0.325 PosECR [0.286,0.365]. Exhaustive QC finds 449/500 development and 465/500 confirmation pairs strict-valid (34 unreadable sources; one target present). Frozen 100-row reannotation reaches 0.900 agreement ($\kappa=0.445$), with all disagreements adjudicated. The frozen 500 remains the prespecified primary estimand; Valid-465 is an exhaustive post-lock QC sensitivity. Filtering changes any confirmation hFPR by at most 0.009 and moves Qwen Target--Full accuracy from +0.003 [$-$0.014,+0.020] to $-$0.003 [$-$0.022,+0.014].

\subsection{Deployment Policy and Realized Cost}

Component ablations confirm a quality--coverage trade-off: stronger Qwen/InternVL region protection raises PosECR but can lower accuracy or raise hFPR, while an InternVL prior sweep changes coverage without consistent AUROC gains.

Position semantics form a separate materialization variable. Preserving rather than compacting logical IDs flips 279/1000 LLaVA development decisions and 101/536 InternVL held-out decisions under the named shared thresholds. With unmatched EasyOCR detections, InternVL Soft evidence retains 0.899 PosECR despite 56.4\% missed positive overlap; LLaVA hard protection reaches 0.657 versus 0.402 without boxes. On 96 open-QA cases, assisted selection scores 0.771 versus 0.703 for question-only Grid and 0.746 for a box-enriched-input control, but remains below Full at 0.999. Position and detector policies therefore belong in the reporting contract.

Table~\ref{tab:validity} measures execution, not FLOPs. Qwen Target (20\%) raises batch-prefill throughput from 12.8 to 55.2 samples/s (4.32$\times$) and lowers incremental peak memory by 76.4\%; LLaVA/InternVL reach 2.24--2.34$\times$, with vision encoding unchanged. Including selection, LLaVA Target/AnchorPrune remain 1.40/1.22$\times$ faster, while SCOPE/CoIn fall below Full. Online EasyOCR erases box-assisted single-sample gains; Qwen's 1.45$\times$ TTFT gain becomes 1.06$\times$ over 32-token generation.

\subsection{Task Scope Changes the Frontier}

Original-question generation requires substantially more evidence than binary target verification. On OCRBench, increasing Qwen retention from 30\% to 70\% raises exact match from 0.480 to 0.660 and ANLS from 0.619 to 0.736, approaching the Full scores of 0.690 and 0.777.

At 70\% retention on full TextVQA/DocVQA validation, Qwen Target+Grid scores 0.795/0.846 versus 0.828/0.940 for Full, and LLaVA Target scores 0.359/0.162 versus 0.485/0.216; all paired intervals exclude zero. The selector trails Random in all four model--task pairs and exceeds Grid only on Qwen DocVQA. On 96 human-corrected cases, raising retention from 30\% to 70\% increases worst-region ECR from 0.216 to 0.613 and the all-regions-covered fraction from 0.105 to 0.729. Thus multi-region generation needs greater, task-specific coverage.

\section{Discussion and Limitations}

Visual-token pruning separates answer behavior, spatial traceability, and realized cost. Same-budget masks differ sharply in PosECR, while merging separates lineage from local anchors. Matched controls and interventions show that neither arbitrary retention nor lexical priors explain the gap, and higher coverage need not reduce answer risk.

Selectors should therefore be chosen on a quality--risk--traceability frontier rather than transferred by keep ratio or accuracy. Qwen Target preserves more annotated support than matched controls, but its Target--Full interval does not establish noninferiority; LLaVA and InternVL require backbone-specific calibration. Deployment must distinguish \emph{box-free}, \emph{existing OCR/layout}, and \emph{online-detector} regimes because detector, position, and task scope change both behavior and cost.

Audit semantics must match compression. In deletion, retained cells provide direct paths, but coarse or dynamic cells can inflate recall, motivating LPP and Geo-F1. In merging, lineage records source participation whereas AnchorECR requires a locally traceable representative; both prevent exhaustive assignment from masquerading as localization. PosECR and NegSRC remain separate because positive support and a negative probe's confusable source have different meanings.

Efficiency requires the same discipline. Keep ratio omits position handling, selection, materialization, vision encoding, and online detection. Shortened prefixes accelerate batched prefill, but TTFT gains are smaller and detector latency can erase them; reports must identify the model, task, batch size, position policy, detector source, and measured path.

Three limits bound the scope. ECR measures geometric availability, not causal use; coarse cells require precision checks, and intervention agreement is strongest for Qwen. Aggressive points concern target verification, whereas open QA needs higher retention. Assisted selectors inherit detector error and latency, and annotation disagreement limits exact context claims. These boundaries restrict transfer of a frontier point, not the observed separation between quality and traceability.

\section{Conclusion}

Visual-token pruning cannot be characterized by answer quality and compression ratio alone: prefixes with similar budgets and accuracy can retain fundamentally different spatial support. Our evidence-risk audit couples answer behavior with source-aware provenance, interventions, and realized cost, revealing model-, calibration-, and task-specific quality--risk--traceability frontiers across three MLLMs. Pruning evaluations should therefore report surviving visual support and realized efficiency under task- and operator-appropriate semantics.

\bibliography{references}

\appendix
\setcounter{table}{0}
\setcounter{figure}{0}
\renewcommand{\thetable}{S\arabic{table}}
\renewcommand{\thefigure}{S\arabic{figure}}

\section{Supplementary Material}

This supplement provides protocol details, extended budget curves, complementary controls, evidence diagnostics, external-method comparisons, detector robustness, open-QA checks, CUDA timing, and transfer boundaries.

\section{Method and Protocol Details}

\textbf{Selector variants.}
Target ranks by the queried target span; Full-prompt uses every non-image prompt token; Target+Grid adds a uniform grid floor. These variants are box-free. Protected hard-reserves supplied OCR/layout regions, whereas Soft evidence adds a region prior; ``risk-constrained'' rows change only the development-selected decision threshold and remain sensitivity analyses.

\textbf{Dataset and lock.}
Each TextOCR-Hard split contains 500 images and paired positive/near-miss probes for one unique small OCR word per image. The source box denotes positive support but only the confusable source region on negatives. The confirmation split was generated with seed 20260720 after excluding every development image. Before inference, we froze Qwen Full/Target/Random/Grid at 30\% and LLaVA Full/Protected/Random/VisionZip at 40\%, including seeds and decoding. SCOPE and Qwen VisionZip were later added at already evaluated budgets without adapting selectors or ratios to confirmation outcomes. Five additional Random masks are explicitly post-lock sensitivity runs. No confirmation result changed an operating point; constructors, digests, and the complete lexical audit are in the artifact.

\textbf{Evaluation contract.}
We score the complete continuations `` yes'' and `` no'' under each chat template and predict yes iff $\ell_{\mathrm{no}}-\ell_{\mathrm{yes}}\geq t$. Their template-expanded lengths match within each backbone. Qwen/LLaVA use $t=0$; InternVL selects thresholds only on 464 development probes and evaluates on 536 image-group-disjoint probes. PosECR covers positive source boxes; NegSRC covers negative confusable-source boxes and is not evidence of target absence. P/N avg.\ is their descriptive macro-average, CenterR tests represented region centers, and PatchR measures represented intersecting source cells. The lock controls operating-point adaptation but is not a preregistered noninferiority or multiplicity-controlled trial, so paired accuracy differences and complementary audit axes are interpreted descriptively.

\textbf{Statistics and controls.}
Paired TextOCR-Hard intervals average the two probes within each image and bootstrap 500 image-level differences 10,000 times; open-QA intervals resample examples. Multi-seed Random sensitivity additionally resamples masks and images. Image-dependence controls keep prompts and labels fixed but use a gray image or three collision-free wrong-image derangements (seeds 101/202/303); paired probes share the same unrelated image. Raw, Unicode, case-folded, whitespace, alphanumeric, and ASCII-folded matching excludes source/target collisions. InternVL retains the Full-derived threshold for every image condition. Cached scores, seeds, and full per-seed outputs are in the artifact.

\textbf{Token-to-region geometry.}
Boxes are normalized to the processed image plane. Qwen derives its post-merge raster from \texttt{image\_grid\_thw}; LLaVA maps 576 tokens to a $24\times24$ grid; InternVL maps local post-downsampling grids through each dynamic tile and treats thumbnail tokens as a full-image grid. Every backend verifies exact agreement between generated boxes and projected visual-token count.

\textbf{Compute environment.}
Runs use one NVIDIA A800 80GB GPU with CUDA 12.4, PyTorch 2.6.0, and Transformers 4.57.1. The artifact records the complete environment, model snapshot/fingerprints, and package lock.

\section{Extended TextOCR-Hard Budget Curves}

Table~\ref{tab:s-budget-frontier} shows representative rows around the main operating points, testing whether the reported settings lie on stable quality--risk--coverage frontiers rather than being isolated ratios. The complete selector and budget sweeps are included in the artifact.

\begin{table*}[!tbp]
\centering
\small
\begin{tabular*}{\textwidth}{@{\extracolsep{\fill}}llrrrrr@{}}
\toprule
Model & Method & $n$ & Keep & Acc. & hFPR & P/N avg. \\
\midrule
Qwen & Full & 1000 & 1.000 & 0.787 & 0.236 & 1.000 \\
Qwen & Target (20\%) & 1000 & 0.200 & 0.793 & 0.224 & 0.524 \\
Qwen & Target (25\%) & 1000 & 0.251 & 0.795 & 0.214 & 0.565 \\
Qwen & Target (30\%) & 1000 & 0.301 & 0.798 & 0.222 & 0.604 \\
\midrule
LLaVA & Full & 1000 & 1.000 & 0.626 & 0.304 & 1.000 \\
LLaVA & Protected (20\%) & 1000 & 0.201 & 0.623 & 0.542 & 1.000 \\
LLaVA & Protected (40\%) & 1000 & 0.401 & 0.661 & 0.298 & 1.000 \\
LLaVA & Protected (50\%) & 1000 & 0.500 & 0.667 & 0.308 & 1.000 \\
\midrule
InternVL & Full & 536 & 1.000 & 0.646 & 0.306 & 1.000 \\
InternVL & Target (40\%) & 536 & 0.400 & 0.642 & 0.396 & 0.825 \\
InternVL & Target (50\%) & 536 & 0.500 & 0.640 & 0.179 & 0.872 \\
InternVL & Soft evidence (50\%) & 536 & 0.500 & 0.653 & 0.328 & 0.893 \\
\bottomrule
\end{tabular*}
\caption{Representative TextOCR-Hard budget rows. Qwen and LLaVA use development data; InternVL uses its image-disjoint held-out split with selector-specific development thresholds. P/N avg.\ averages positive-support and negative-source geometric coverage. Trends are interpreted within a backbone: stronger protection raises coverage but does not monotonically improve answer risk.}
\label{tab:s-budget-frontier}
\end{table*}

\section{Selector Hyperparameter Sensitivity}

Table~\ref{tab:s-selector-sensitivity} summarizes one-factor sweeps at fixed budgets; complete row-level results are included in the artifact. Qwen uses the locked confirmation set at 30\% retention. Increasing the norm contribution raises source coverage but eventually damages answer behavior; top-two query aggregation is stronger in accuracy than top-one or top-four. Grid reservation does not dominate pure Target on this single-region protocol, so Target+Grid remains an optional multi-region hedge. InternVL uses the common Full-derived threshold at 50\% retention: increasing $\beta$ raises source coverage from 0.872 to 0.940, while AUROC and hFPR remain non-monotonic. The frozen settings are therefore compromise points, not uniquely optimal scalars.

\begin{table*}[!tbp]
\centering
\small
\begin{tabular*}{\textwidth}{@{\extracolsep{\fill}}llrrrrrl@{}}
\toprule
Model & Sweep & Settings & Acc. range & hFPR range & AUROC range & Coverage range & Frozen \\
\midrule
Qwen & relevance weight & 0.50--1.00 & 0.757--0.788 & 0.232--0.294 & -- & 0.544--0.611 & 0.85 \\
Qwen & query top-$k$ & 1, 2, 4 & 0.774--0.786 & 0.240--0.260 & -- & 0.553--0.571 & 2 \\
Qwen & grid reservation & 0.00--0.75 & 0.784--0.786 & 0.238--0.246 & -- & 0.507--0.564 & 0.00 \\
InternVL & evidence weight $\beta$ & 0.00--0.20 & 0.604--0.618 & 0.522--0.556 & 0.688--0.697 & 0.872--0.940 & 0.05 \\
\bottomrule
\end{tabular*}
\caption{Compact selector sensitivity summary. Qwen reports fixed-$t=0$ accuracy and hFPR; InternVL reports common-Full-threshold accuracy/hFPR and threshold-free AUROC. Coverage is P/N avg. Full row-level sweeps, including CenterR and selector-specific thresholds, are provided in the artifact.}
\label{tab:s-selector-sensitivity}
\end{table*}

\section{Same-Budget and Paired Controls}

The same-budget controls ask whether any mask of the same size would suffice. Random and grid controls appear in the main table; Table~\ref{tab:s-same-budget} adds the complementary shuffled-score controls. Table~\ref{tab:s-image-cluster} gives the paired image-cluster intervals, and Table~\ref{tab:s-correct-positive-provenance} conditions on correctly answered positives to measure how often retained token cells still have low or zero overlap with the annotated support region.

\begin{table*}[!tbp]
\centering
\small
\begin{tabular*}{\textwidth}{@{\extracolsep{\fill}}llrrrrrr@{}}
\toprule
Group & Method & $n$ & Acc. & hFPR & Keep & P/N avg. & $\Delta$ avg. \\
\midrule
Qwen (20\%) & Target & 1000 & 0.793 & 0.224 & 0.200 & 0.524 & -- \\
Qwen (20\%) & Shuffled score & 1000 & 0.718 & 0.122 & 0.200 & 0.192 & -0.332 \\
\midrule
Qwen (30\%) & Target & 1000 & 0.798 & 0.222 & 0.301 & 0.604 & -- \\
Qwen (30\%) & Shuffled score & 1000 & 0.739 & 0.158 & 0.301 & 0.288 & -0.317 \\
\midrule
LLaVA (40\%) & Protected & 1000 & 0.661 & 0.298 & 0.401 & 1.000 & -- \\
LLaVA (40\%) & Shuffled score & 1000 & 0.653 & 0.278 & 0.401 & 0.412 & -0.588 \\
\midrule
InternVL (50\%) & Soft evidence & 536 & 0.653 & 0.328 & 0.500 & 0.893 & -- \\
InternVL (50\%) & Shuffled score & 536 & 0.634 & 0.183 & 0.500 & 0.780 & -0.113 \\
\bottomrule
\end{tabular*}
\caption{Complementary shuffled-score controls at the same visual-token budgets. P/N avg.\ is the geometric macro-average over positive support and negative source regions; the InternVL rows use selector-specific accuracy-optimal development thresholds and are diagnostic rather than the shared-threshold primary comparison.}
\label{tab:s-same-budget}
\end{table*}

\begin{table*}[!tbp]
\centering
\small
\begin{tabular*}{\textwidth}{@{\extracolsep{\fill}}p{0.28\textwidth}lrrrrr@{}}
\toprule
Comparison & Model & $n_{\mathrm{img}}$ & $\Delta$Acc. & Acc. CI & $\Delta$hFPR & hFPR CI \\
\midrule
Target (30\%) -- Full, confirmation & Qwen & 500 & +0.003 & [-0.014,+0.020] & -0.008 & [-0.038,+0.022] \\
Target (30\%) -- Random (30\%), confirmation & Qwen & 500 & +0.047 & [+0.021,+0.073] & +0.080 & [+0.044,+0.118] \\
Target (30\%) -- Grid (30\%), confirmation & Qwen & 500 & +0.063 & [+0.039,+0.087] & +0.098 & [+0.060,+0.134] \\
Target (30\%) -- VisionZip (30\%), confirmation & Qwen & 500 & +0.038 & [+0.018,+0.059] & +0.042 & [+0.010,+0.076] \\
Protected (40\%) -- Full, confirmation & LLaVA & 500 & +0.018 & [-0.006,+0.041] & +0.036 & [-0.006,+0.076] \\
Protected (40\%) -- VisionZip, confirmation & LLaVA & 500 & +0.061 & [+0.034,+0.088] & -0.318 & [-0.366,-0.272] \\
CoIn (40\%) -- SCOPE, confirmation & LLaVA & 500 & -0.007 & [-0.026,+0.013] & +0.112 & [+0.076,+0.148] \\
CoIn (40\%) -- Target, confirmation & LLaVA & 500 & -0.045 & [-0.071,-0.019] & +0.100 & [+0.054,+0.146] \\
SCOPE pure coverage -- SCOPE, confirmation & LLaVA & 500 & -0.007 & [-0.024,+0.010] & +0.086 & [+0.052,+0.120] \\
Soft evidence (50\%) -- Full & InternVL & 268 & +0.007 & [-0.032,+0.047] & +0.022 & [-0.041,+0.086] \\
Soft evidence (50\%) -- Grid (50\%) & InternVL & 268 & +0.006 & [-0.032,+0.043] & +0.075 & [+0.019,+0.134] \\
\bottomrule
\end{tabular*}
\caption{Paired image-cluster bootstrap on TextOCR-Hard (10,000 draws). Rows explicitly marked as confirmation results use the locked confirmation set; unmarked Qwen/LLaVA diagnostics use development data, and InternVL uses its held-out split. Lower hFPR is better.}
\label{tab:s-image-cluster}
\end{table*}

\begin{table*}[!tbp]
\centering
\small
\begin{tabular*}{\textwidth}{@{\extracolsep{\fill}}lrrrrr@{}}
\toprule
Method & Correct positives & Lin. PosECR & AnchorECR & PosECR $<0.50$ & PosECR $=0$ \\
\midrule
Qwen Target (30\%) & 406 & 0.672 & 0.672 & 27.8 [23.7,32.4] & 13.1 [10.1,16.7] \\
Qwen Random (30\%) & 319 & 0.289 & 0.289 & 71.2 [66.0,75.9] & 37.0 [31.9,42.4] \\
Qwen Grid (30\%) & 294 & 0.371 & 0.371 & 65.3 [59.7,70.5] & 20.7 [16.5,25.7] \\
\midrule
LLaVA Protected (40\%) & 319 & 1.000 & 1.000 & 0.0 [0.0,1.2] & 0.0 [0.0,1.2] \\
LLaVA Random (40\%) & 297 & 0.374 & 0.374 & 63.6 [58.0,68.9] & 31.0 [26.0,36.5] \\
LLaVA Target (40\%) & 411 & 0.486 & 0.486 & 52.1 [47.2,56.9] & 27.0 [22.9,31.5] \\
LLaVA SCOPE (40\%) & 367 & 0.637 & 0.637 & 34.3 [29.7,39.3] & 16.9 [13.4,21.1] \\
LLaVA AnchorPrune (40\%) & 371 & 0.650 & 0.650 & 33.2 [28.6,38.1] & 18.1 [14.5,22.3] \\
LLaVA SCOPE pure coverage (40\%) & 403 & 0.626 & 0.626 & 35.5 [31.0,40.3] & 15.6 [12.4,19.5] \\
LLaVA CoIn (40\%) & 416 & 0.677 & 0.677 & 31.0 [26.8,35.6] & 14.2 [11.2,17.9] \\
LLaVA VisionZip (40\%) & 417 & 1.000 & 0.576 & 0.0 [0.0,0.9] & 0.0 [0.0,0.9] \\
\bottomrule
\end{tabular*}
\caption{Correct-positive spatial-provenance audit on locked confirmation. The last two columns use lineage PosECR and report percentages with Wilson 95\% intervals; Correct positives gives each denominator. For non-merging methods AnchorECR equals lineage PosECR. VisionZip's full lineage but lower AnchorECR separates source participation in a merge from locally attributable output positions.}
\label{tab:s-correct-positive-provenance}
\end{table*}

\section{Statistical and Construction Audits}

TextOCR-Hard contains paired positive and negative probes from the same image, so Table~\ref{tab:s-image-cluster} uses an image-cluster bootstrap. On locked confirmation, Qwen Target--Full remains centered near zero. Against matched Random and Grid, its PosECR gains are +0.350 (95\% CI [+0.304,+0.394]) and +0.302 ([+0.259,+0.346]); against native VisionZip, Target gains +0.038 accuracy ([+0.018,+0.059]) but loses 0.380 lineage PosECR ([-0.414,-0.348]). VisionZip's exhaustive source lineage covers every region by construction, whereas its representative output anchors cover only 0.846 PosECR for Qwen and 0.563 for LLaVA; Table~\ref{tab:s-correct-positive-provenance} gives the corresponding correctly answered-positive audit. LLaVA protected pruning remains stronger than the VisionZip adaptation in answer behavior, and InternVL soft evidence improves geometric coverage without a symmetric-calibration hFPR gain.

Table~\ref{tab:s-hard-negative-quality} separates development and confirmation construction audits. Neither split has a collision under raw, Unicode, case-folded, or whitespace-stripped matching; one confirmation pair collides only after punctuation-stripping alphanumeric normalization. Table~\ref{tab:s-hard-negative-human-qc} reports exhaustive human audits of all 500 negatives in \emph{each} split. Development QC confirms target absence for 486 and strict source/box validity for 449; filtering changes any hFPR by at most 0.013. Locked-confirmation QC confirms 465 strict-valid negatives, while 34 have an unreadable source and one contains the queried target. Its frozen 100-row independent reannotation reaches 0.900 binary-validity agreement ($\kappa=0.445$), with all 10 disagreements adjudicated. Filtering to the 465 valid image pairs changes any confirmation hFPR by at most 0.009. The prespecified 500-image table remains the primary readout; on the post-QC subset, Qwen Target--Full accuracy is $-$0.003 (95\% CI [$-$0.022,+0.014]), versus +0.003 ([$-$0.014,+0.020]) originally.

The main paper summarizes matched, blank, and three-seed wrong-image controls. On the stricter lexically plausible subset, matched versus mismatch accuracy is 0.672 versus 0.497 for Qwen ($n=128$), 0.609 versus 0.544 for LLaVA ($n=128$), and 0.712 versus 0.480 for InternVL ($n=66$). Full per-seed image-dependence results are included in the artifact. The random baselines are also stable over deterministic masks: on locked Qwen confirmation, the prespecified row plus five fixed post-lock replicates give accuracy $0.733{\pm}0.005$ (0.726--0.739), hFPR $0.163{\pm}0.004$ (0.158--0.168), and PosECR $0.294{\pm}0.016$ (0.270--0.315). A two-stage seed-and-image bootstrap gives Target--Random +0.053 accuracy [0.033,0.073] and +0.325 PosECR [0.286,0.365].

ECR throughout the audit is geometric spatial provenance. A token's cell identifies its grid origin, not the full information encoded after contextual vision processing. Therefore zero ECR means that no retained token cell overlaps the annotated region; it does not prove that all information about that region has been erased from every retained embedding.

Tables~\ref{tab:s-confirmation-conditional-construct} and \ref{tab:s-construct-validity} directly test this construct boundary on positive probes. The first joins Target, Random, Grid, and VisionZip masks at a fixed 30\% Qwen confirmation budget. ECR remains associated with selected-minus-Full yes-margin after measured controls and under image--method fixed effects, so the relation is not explained only by image difficulty or selector identity; excluding merge-based VisionZip preserves the association. The second uses complementary development-set interventions. For retained cells that touch an annotated region, local provenance precision (LPP) divides their in-region union area by their total union area; it therefore penalizes a coarse cell that covers a small word box. Geo-F1 is the harmonic mean of ECR and LPP. ECR is positively associated with three Qwen interventions, but the LLaVA and InternVL intervals include zero. InternVL additionally has a restricted ECR range under Soft evidence. A scale-stratified audit confirms that coarse geometry can inflate availability: in InternVL's largest-cell tertile, ECR is 0.769 while LPP is only 0.094. These results support convergent validity on Qwen and a geometry audit on all models, not a backbone-invariant causal metric. We therefore interpret ECR within a backbone, policy, budget, and token grid and do not rank backbones by absolute ECR.

InternVL requires an explicit yes/no operating point because its default likelihood threshold is strongly yes-biased on TextOCR-Hard. We predict yes iff $\ell_{\mathrm{no}}-\ell_{\mathrm{yes}}\geq t$, with thresholds selected only on the 464-probe development split and all reported values measured on the image-group-disjoint 536-probe held-out split. The primary main-paper comparison applies the Full-derived threshold $t=2.043$ unchanged to every selector and reports threshold-free AUROC. Table~\ref{tab:s-internvl-operating} also separates selector-specific accuracy-optimal thresholds, under which Soft evidence uses $t=2.344$, from this common-threshold comparison. Soft evidence changes AUROC only from 0.687 to 0.696 and has 0.530 hFPR under the shared threshold. A conservative risk-constrained threshold $t=2.646$ lowers its hFPR to 0.175 without changing the mask, raw margins, PosECR, or NegSRC, but flips 85 test predictions from yes to no and none in the reverse direction. We therefore treat the own-threshold and 0.175 rows as operating-point sensitivity results; the selector's clearer contribution is 0.902 PosECR and 0.883 NegSRC with a small ranking improvement.

Table~\ref{tab:s-qwen-llava-operating} provides the symmetric diagnostic for Qwen and LLaVA. For each backbone, one threshold is selected from its independent Full-prefix development scores and applied unchanged to every locked-confirmation selector; AUROC remains threshold-free. Qwen Target nearly matches Full in AUROC (0.855 vs.\ 0.856), while matched Random, Grid, and VisionZip are lower. LLaVA Target has higher AUROC than Full (0.713 vs.\ 0.666), but its hFPR remains 0.442 under the shared threshold. Its fixed-threshold behavior therefore includes a substantial calibration shift, but is not explained by calibration alone.

\begin{table*}[!tbp]
\centering
\small
\begin{tabular*}{\textwidth}{@{\extracolsep{\fill}}lrr@{}}
\toprule
Construction or lexical audit & Development & Locked confirmation \\
\midrule
Negative probes / images & 500 / 500 & 500 / 500 \\
Mean edit distance & 1.012 & 1.012 \\
Mean normalized edit distance & 0.208 & 0.208 \\
Same-image collision under any normalizer & 0/500 & 1/500 \\
Source--target collapse under any normalizer & 0/500 & 0/500 \\
Edit distance greater than 2 after NFKC--casefold & 0/500 & 0/500 \\
Other-image token matches after NFKC--casefold & 47/500 & 47/500 \\
Other-image token matches after alphanumeric filtering & 58/500 & 64/500 \\
Substitution / deletion / other edits & 365 / 129 / 6 & 376 / 118 / 6 \\
\bottomrule
\end{tabular*}
\caption{Hard-negative construction and lexical-normalization audits by split. Normalizers include raw trimming, NFC, NFKC, case folding, whitespace removal, alphanumeric filtering, and ASCII folding. The single confirmation collision occurs only after punctuation-stripping alphanumeric normalization (\texttt{in'} versus a separate \texttt{in} token); raw and Unicode-normalized forms do not collide. Other-image matches indicate plausible OCR strings elsewhere in TextOCR, not same-image label errors.}
\label{tab:s-hard-negative-quality}
\end{table*}

Human QC reviews all 500 paired source words in each split. On development, 486 targets are confirmed absent and 449 pairs meet strict source-visible, target-absent, and box-matched validity; filtering changes any of 17 methods' hFPR by at most 0.013. On locked confirmation, 465 pairs meet the same strict criteria, 34 sources are unreadable, and one queried target is present. Because the positive probe queries this same source word and uses the same box, source visibility and box--text agreement also audit positive-probe readability and localization rather than only negative construction. A frozen 100-row independent reannotation gives 0.900 binary-validity agreement ($\kappa=0.445$), and all 10 disagreements are adjudicated.

\begin{table*}[!tbp]
\centering
\small
\begin{tabular*}{\textwidth}{@{\extracolsep{\fill}}llrrrrrr@{}}
\toprule
Model & Method & Acc. & hFPR & AUROC & PosECR & AncECR & NegSRC \\
\midrule
Qwen & Full & 0.801 & 0.249 & 0.877 & 1.000 & 1.000 & 1.000 \\
Qwen & Target (30\%) & 0.798 & 0.245 & 0.875 & 0.634 & 0.634 & 0.514 \\
Qwen & Random (30\%) & 0.752 & 0.161 & 0.792 & 0.270 & 0.270 & 0.298 \\
Qwen & Grid (30\%) & 0.735 & 0.144 & 0.765 & 0.321 & 0.321 & 0.321 \\
Qwen & VisionZip (30\%) & 0.763 & 0.202 & 0.807 & 1.000 & 0.851 & 1.000 \\
\midrule
LLaVA & Full & 0.631 & 0.320 & 0.675 & 1.000 & 1.000 & 1.000 \\
LLaVA & Protected (40\%) & 0.651 & 0.353 & 0.704 & 1.000 & 1.000 & 1.000 \\
LLaVA & Random (40\%) & 0.674 & 0.260 & 0.719 & 0.375 & 0.375 & 0.388 \\
LLaVA & Target (40\%) & 0.625 & 0.583 & 0.720 & 0.479 & 0.479 & 0.448 \\
LLaVA & SCOPE (40\%) & 0.590 & 0.561 & 0.652 & 0.613 & 0.613 & 0.613 \\
LLaVA & AnchorPrune (40\%) & 0.584 & 0.583 & 0.650 & 0.636 & 0.636 & 0.641 \\
LLaVA & CoIn (40\%) & 0.584 & 0.669 & 0.672 & 0.672 & 0.672 & 0.688 \\
LLaVA & VisionZip (40\%) & 0.586 & 0.669 & 0.672 & 1.000 & 0.562 & 1.000 \\
\bottomrule
\end{tabular*}
\caption{Complete post-QC readout on the 465 human-valid locked-confirmation image pairs. The original 500-image table remains the prespecified primary analysis. Across rows, filtering changes hFPR by at most 0.009, AUROC by 0.022, PosECR/AncECR by 0.014, and NegSRC by 0.007. Qwen Target--Full accuracy changes from +0.003 [$-0.014,+0.020$] to $-0.003$ [$-0.022,+0.014$] under image-cluster bootstrap.}
\label{tab:s-hard-negative-human-qc}
\end{table*}

\begin{table*}[!tbp]
\centering
\small
\begin{tabular*}{\textwidth}{@{\extracolsep{\fill}}llrrr@{}}
\toprule
Specification & Predictor & $n$ images & $r_s$ & 95\% CI \\
\midrule
Measured controls + method FE & ECR & 500 & 0.232 & [0.191,0.273] \\
Measured controls + method FE & LPP & 500 & 0.229 & [0.184,0.273] \\
Measured controls + method FE & Geo-F1 & 500 & 0.240 & [0.196,0.283] \\
Image + method FE & ECR & 500 & 0.237 & [0.186,0.286] \\
Image + method FE & LPP & 500 & 0.219 & [0.167,0.271] \\
Image + method FE & Geo-F1 & 500 & 0.232 & [0.181,0.281] \\
\midrule
Deletion only, measured controls + method FE & ECR & 500 & 0.279 & [0.230,0.328] \\
Deletion only, image + method FE & ECR & 500 & 0.253 & [0.188,0.315] \\
\bottomrule
\end{tabular*}
\caption{Conditional construct audit over Qwen masks at the same 30\% locked-confirmation budget. The first six rows use 2,000 method--image observations from Target, Random, Grid, and VisionZip; deletion-only sensitivity excludes VisionZip and uses 1,500. The outcome is selected-minus-Full yes-margin on positive probes. Measured controls are ranked log evidence-box area, ranked log median token-cell area, Full margin, and method indicators; keep ratio is fixed by design. The second specification absorbs all image- and method-level shifts. Intervals use 10,000 image-cluster bootstrap draws retaining all masks in each scope. These associations support convergent validity, not causal use.}
\label{tab:s-confirmation-conditional-construct}
\end{table*}

\begin{table*}[!tbp]
\centering
\small
\begin{tabular*}{\textwidth}{@{\extracolsep{\fill}}lrrrrrrrrr@{}}
\toprule
Model & $n$ & ECR & LPP & Geo-F1 & Occ. $n$ & Occ. $r_s$ & Partial Occ. & Delete $r_s$ & Partial Delete \\
\midrule
Qwen3-VL-8B & 500 & 0.651 & 0.205 & 0.295 & 100 & 0.334 & 0.102 & 0.308 & 0.288 \\
LLaVA-1.5-7B & 500 & 0.458 & 0.112 & 0.171 & 50 & 0.270 & 0.275 & -- & -- \\
InternVL3.5-8B & 500 & 0.916 & 0.166 & 0.255 & 50 & 0.158 & 0.159 & 0.029 & 0.011 \\
\bottomrule
\end{tabular*}
\caption{Positive-probe construct-validity audit. Occ. correlates ECR with the evidence-box yes-margin drop after subtracting a same-area random-mask drop; Delete uses the selected-prefix minus evidence-deleted margin. Partial Spearman controls ranked log evidence-box area, log median token-cell area, and Full-prefix yes-margin; LLaVA's constant cell area is dropped. Marginal 95\% intervals are Qwen Occ. [0.138,0.511], Qwen Delete [0.227,0.389], LLaVA Occ. [$-$0.012,0.519], InternVL Occ. [$-$0.147,0.434], and InternVL Delete [$-$0.058,0.112]. Partial intervals are [$-$0.082,0.285], [0.203,0.371], [$-$0.025,0.535], [$-$0.119,0.414], and [$-$0.075,0.096], respectively, from 10,000 paired bootstrap resamples.}
\label{tab:s-construct-validity}
\end{table*}

\begin{table*}[!tbp]
\centering
\small
\begin{tabular*}{\textwidth}{@{\extracolsep{\fill}}lrrrrrr@{}}
\toprule
InternVL row & AUROC & Own Acc. & Own hFPR & Shared Acc. & Shared hFPR & P/N avg. \\
\midrule
Full & 0.687 & 0.646 & 0.306 & 0.646 & 0.306 & 1.000 \\
Target (50\%) & 0.695 & 0.640 & 0.179 & 0.618 & 0.522 & 0.872 \\
Grid (50\%) & 0.699 & 0.647 & 0.254 & 0.623 & 0.474 & 0.695 \\
Random (50\%) & 0.677 & 0.614 & 0.336 & 0.604 & 0.485 & 0.742 \\
Soft evidence (50\%) & 0.696 & 0.653 & 0.328 & 0.616 & 0.530 & 0.893 \\
\bottomrule
\end{tabular*}
\caption{InternVL selector--calibration decomposition on the 536-probe test split. Own columns use each selector's accuracy-optimal development threshold; Shared columns apply the Full-derived threshold $t=2.043$ to every row. AUROC is threshold-free.}
\label{tab:s-internvl-operating}
\end{table*}

\begin{table*}[!tbp]
\centering
\small
\begin{tabular*}{\textwidth}{@{\extracolsep{\fill}}llrrrrr@{}}
\toprule
Model & Method & AUROC & Shared Acc. & Shared hFPR & $t{=}0$ Acc. & $t{=}0$ hFPR \\
\midrule
Qwen & Full & 0.856 & 0.812 & 0.132 & 0.783 & 0.248 \\
Qwen & Target (30\%) & 0.855 & 0.810 & 0.142 & 0.786 & 0.240 \\
Qwen & Random (30\%) & 0.776 & 0.757 & 0.072 & 0.739 & 0.160 \\
Qwen & Grid (30\%) & 0.755 & 0.723 & 0.080 & 0.723 & 0.142 \\
Qwen & VisionZip (30\%) & 0.788 & 0.764 & 0.110 & 0.748 & 0.198 \\
\midrule
LLaVA & Full & 0.666 & 0.623 & 0.220 & 0.625 & 0.316 \\
LLaVA & Protected (40\%) & 0.694 & 0.653 & 0.224 & 0.643 & 0.352 \\
LLaVA & Random (40\%) & 0.712 & 0.657 & 0.150 & 0.669 & 0.256 \\
LLaVA & Target (40\%) & 0.713 & 0.653 & 0.442 & 0.624 & 0.574 \\
LLaVA & SCOPE (40\%) & 0.644 & 0.593 & 0.454 & 0.586 & 0.562 \\
LLaVA & AnchorPrune (40\%) & 0.642 & 0.596 & 0.474 & 0.578 & 0.586 \\
LLaVA & CoIn (40\%) & 0.664 & 0.595 & 0.562 & 0.579 & 0.674 \\
\bottomrule
\end{tabular*}
\caption{Qwen/LLaVA selector--calibration decomposition on locked confirmation. Shared columns apply one threshold selected from the corresponding Full-prefix development scores to every selector: $t=1.062$ for Qwen and $t=0.088$ for LLaVA. The primary results retain the prespecified $t=0$ rule; this post-lock diagnostic separates threshold-free ranking from operating-point shift.}
\label{tab:s-qwen-llava-operating}
\end{table*}

\section{Evidence Controls and Diagnostics}

The diagnostic rows test whether annotated regions affect behavior. On positives, evidence-only and anti-evidence prefixes isolate or remove answer-supporting OCR evidence. On negatives, the same operations isolate or remove the confusable source word and are interpreted only as source-region interventions. Table~\ref{tab:s-evidence-controls} includes the Qwen deletion--restoration contrasts needed for the main claim; the artifact records its full 25/50/100\% curve. In the separate InternVL diagnostic-mask check, removing selected evidence lowers accuracy from 0.647 to 0.586; complete evidence restoration returns it to 0.647, whereas matched random restoration reaches only 0.597. Table~\ref{tab:s-diagnostics} reports complementary token-level logit-drop and image-region occlusion interventions across all three backbones. Table~\ref{tab:s-text-replacement-boundary} adds a stricter semantic edit: a local TextOCR word is replaced by its paired near-miss string, while aligned sham and erase controls separate semantic replacement from rendering artifacts. Human review accepts 102 of 106 automatically screened edits. For these 102 cases, Table~\ref{tab:s-edit-region} derives the substituted-glyph box from the same renderer and audits 30\% Qwen masks against the localized edit rather than the full word. Qwen responds consistently on a meaningful subset, whereas InternVL rarely satisfies the same controls. These diagnostics support model-dependent evidence use on positives and source sensitivity on negatives, rather than universal causal proof.

\begin{table*}[!tbp]
\centering
\small
\begin{tabular*}{\textwidth}{@{\extracolsep{\fill}}llrrrrrr@{}}
\toprule
Model & Prefix & $n$ & Acc. & hFPR & Keep & P/N avg. & Note \\
\midrule
Qwen & Full & 1000 & 0.787 & 0.236 & 1.000 & 1.000 & full-token reference \\
Qwen & Target (20\%) & 1000 & 0.793 & 0.224 & 0.200 & 0.524 & main efficiency point \\
Qwen & Evidence only (20\%) & 1000 & 0.771 & 0.242 & 0.200 & 1.000 & evidence sufficiency \\
Qwen & Anti-evidence (20\%) & 1000 & 0.672 & 0.078 & 0.200 & 0.000 & evidence removal \\
Qwen & Target (30\%) & 1000 & 0.798 & 0.222 & 0.301 & 0.604 & main quality point \\
Qwen & Evidence only (30\%) & 1000 & 0.778 & 0.234 & 0.301 & 1.000 & evidence sufficiency \\
Qwen & Anti-evidence (30\%) & 1000 & 0.693 & 0.100 & 0.301 & 0.000 & evidence removal \\
\midrule
Qwen & Del.-rest. selected & 1000 & 0.798 & 0.218 & 0.301 & 0.604 & restoration reference \\
Qwen & Del.-rest. removed & 1000 & 0.769 & 0.198 & 0.299 & 0.000 & evidence removed \\
Qwen & Restore 25\% evidence & 1000 & 0.787 & 0.218 & 0.299 & 0.357 & 62.1\% recovery \\
Qwen & Restore 25\% random & 1000 & 0.772 & 0.196 & 0.299 & 0.000 & 10.3\% recovery \\
\midrule
LLaVA & Full & 1000 & 0.626 & 0.304 & 1.000 & 1.000 & full-token reference \\
LLaVA & Protected (40\%) & 1000 & 0.661 & 0.298 & 0.401 & 1.000 & main protected point \\
LLaVA & Evidence only (40\%) & 1000 & 0.639 & 0.238 & 0.401 & 1.000 & evidence sufficiency \\
LLaVA & Anti-evidence (40\%) & 1000 & 0.640 & 0.204 & 0.401 & 0.000 & evidence removal \\
\midrule
InternVL & Full & 536 & 0.646 & 0.306 & 1.000 & 1.000 & calibrated full \\
InternVL & Soft evidence (50\%, risk-constrained) & 536 & 0.647 & 0.175 & 0.500 & 0.893 & auxiliary operating point \\
InternVL & Evidence only (50\%) & 536 & 0.694 & 0.310 & 0.500 & 1.000 & evidence sufficiency \\
InternVL & Anti-evidence (50\%) & 536 & 0.562 & 0.545 & 0.500 & 0.000 & evidence removal \\
\bottomrule
\end{tabular*}
\caption{Region isolation, removal, and representative Qwen restoration controls. On positives, annotated boxes contain answer support; on negatives, they contain only the confusable source word. These oracle interventions are diagnostics rather than deployable selectors. InternVL uses one fixed risk-constrained threshold throughout.}
\label{tab:s-evidence-controls}
\end{table*}

\begin{table*}[!tbp]
\centering
\small
\begin{tabular*}{\textwidth}{@{\extracolsep{\fill}}lrrrrrrr@{}}
\toprule
Model & $n$ & Orig. pair & Repl. pair & Sham pair & Erase pair & Sem. switch & All controls \\
\midrule
Qwen3-VL-8B & 102 & 0.814 & 0.333 & 0.853 & 0.853 & 0.265 & 0.255 \\
InternVL3.5-8B & 102 & 0.402 & 0.176 & 0.333 & 0.637 & 0.049 & 0.029 \\
\bottomrule
\end{tabular*}
\caption{Human-verified semantic text-replacement audit. Of 106 automatically screened original/replacement/sham/erase image sets, human review accepts 102 as valid semantic edits (two invalid and two unclear). ``Sem. switch'' requires both the original source/target pair and replacement source/target pair to be correct; ``All controls'' additionally requires the sham pair to preserve the original reading and the erase pair to reject both strings. Wilson 95\% intervals for the strict rate are [0.180, 0.347] for Qwen and [0.010, 0.083] for InternVL; their paired difference is significant (exact McNemar $p=2.38\times10^{-7}$). The result is human-verified but model-dependent, not a backbone-uniform causal guarantee.}
\label{tab:s-text-replacement-boundary}
\end{table*}

\begin{table*}[!tbp]
\centering
\small
\begin{tabular*}{\textwidth}{@{\extracolsep{\fill}}lrrrrrr@{}}
\toprule
Method & $n$ & Pos. acc. & Word ECR & Edit ECR & Edit ECR$=0$ & Word$\geq0.5$, Edit$<0.5$ \\
\midrule
Full & 102 & 0.873 & 1.000 & 1.000 & 0.000 & 0.000 \\
Target (30\%) & 102 & 0.912 & 0.869 & 0.897 & 0.029 & 0.059 \\
Random (30\%) & 102 & 0.716 & 0.304 & 0.311 & 0.578 & 0.029 \\
Grid (30\%) & 102 & 0.588 & 0.386 & 0.411 & 0.363 & 0.088 \\
\bottomrule
\end{tabular*}
\caption{Human-validated single-character edit-region audit on Qwen. Each positive query asks for the rendered near-miss replacement, so Pos.\ acc.\ is target-present accuracy rather than balanced binary accuracy. The audit box is reconstructed from the renderer's font, size, placement, and unique substitution index; it occupies 19.1\% of the original word box on average and is never supplied to selection. These 102 rows audit localization on human-valid renders; the stricter four-control behavioral test in Table~\ref{tab:s-text-replacement-boundary} succeeds on 26 Qwen rows. Target--Random/Grid paired edit-ECR differences are +0.586 [0.492,0.676] and +0.486 [0.397,0.575].}
\label{tab:s-edit-region}
\end{table*}

\begin{table*}[!tbp]
\centering
\small
\begin{tabular*}{\textwidth}{@{\extracolsep{\fill}}lllrrrr@{}}
\toprule
Model & Comparison & Diagnostic & Overall & Positive & Negative & $n$ \\
\midrule
Qwen & evidence-kept -- removed & token logit-drop & +0.884 & +2.909 & -1.140 & 1000 \\
Qwen & selected prefix -- removed & token logit-drop & +1.089 & +3.316 & -1.139 & 1000 \\
Qwen & selected prefix -- Full & token logit-drop & +0.006 & -0.017 & +0.029 & 1000 \\
Qwen & orig -- evidence-masked & bbox occlusion & +2.190 & +6.261 & -1.882 & 200 \\
Qwen & orig -- random-masked & bbox occlusion & +0.037 & +0.119 & -0.045 & 200 \\
\midrule
LLaVA & evidence-kept -- removed & token logit-drop & +0.005 & +0.039 & -0.029 & 1000 \\
LLaVA & selected prefix -- Full & token logit-drop & +0.019 & +0.053 & -0.015 & 1000 \\
LLaVA & orig -- evidence-masked & bbox occlusion & +0.009 & +0.025 & -0.007 & 100 \\
\midrule
InternVL & evidence-kept -- removed & token logit-drop & +0.218 & +0.146 & +0.291 & 536 \\
InternVL & selected prefix -- Full & token logit-drop & -0.032 & -0.285 & +0.222 & 536 \\
InternVL & orig -- evidence-masked & bbox occlusion & +0.133 & +0.786 & -0.521 & 100 \\
\bottomrule
\end{tabular*}
\caption{Full diagnostic summary. Entries are yes-support margin differences. Positive and negative columns split probes by label; larger positive drops on positives indicate that the retained or occluded region carries answer-relevant evidence.}
\label{tab:s-diagnostics}
\end{table*}

\section{External Method Budget Curves}

\textbf{Coverage ablation.}
Table~\ref{tab:s-llava-official} compares LLaVA external pruning methods under the same TextOCR-Hard likelihood-scoring interface, keep-ratio budget, and yes/no decision rule. SCOPE uses the public LLaVA algorithm at its default $\alpha=1$; an exact-index parity test against the pinned upstream selector passes on four independent tensor cases. Setting its saliency exponent to $\alpha=0$ makes every saliency weight one and isolates SCOPE's facility-location feature-coverage term without changing the 40\% budget or materialization path. On confirmation, pure coverage has unresolved differences from default SCOPE in accuracy ($-0.007$, image-cluster 95\% CI $[-0.024,0.010]$) and PosECR ($+0.005$, $[-0.038,0.047]$), while hFPR increases by 0.086 ($[0.052,0.120]$; exact McNemar $p=8.9\times10^{-7}$). Under our normalized sequence-likelihood readout, 15-bin ECE changes from 0.037 to 0.050, whereas Brier score changes from 0.237 to 0.236. These descriptive calibration metrics use continuation likelihoods rather than the concurrent paper's first-token protocol; they show why feature coverage, calibration, spatial provenance, and answer risk are not interchangeable.

\begin{table*}[!tbp]
\centering
\small
\begin{tabular*}{\textwidth}{@{\extracolsep{\fill}}llrrrrrrr@{}}
\toprule
Method & Ratio & Acc. & hFPR & Pos. acc. & Neg. acc. & Keep & P/N avg. & Anchor avg. \\
\midrule
Protected & 0.20 & 0.623 & 0.542 & 0.788 & 0.458 & 0.201 & 1.000 & 1.000 \\
VisionZip adaptation & 0.20 & 0.554 & 0.780 & 0.888 & 0.220 & 0.201 & 1.000 & 0.336 \\
FastV adaptation & 0.20 & 0.500 & 0.000 & 0.000 & 1.000 & 0.201 & 0.270 & 0.270 \\
\midrule
Protected & 0.30 & 0.627 & 0.440 & 0.694 & 0.560 & 0.300 & 1.000 & 1.000 \\
VisionZip adaptation & 0.30 & 0.555 & 0.734 & 0.844 & 0.266 & 0.300 & 1.000 & 0.466 \\
FastV adaptation & 0.30 & 0.500 & 0.000 & 0.000 & 1.000 & 0.300 & 0.376 & 0.376 \\
\midrule
Protected & 0.40 & 0.661 & 0.298 & 0.620 & 0.702 & 0.401 & 1.000 & 1.000 \\
SCOPE implementation & 0.40 & 0.598 & 0.508 & 0.704 & 0.492 & 0.401 & 0.642 & 0.642 \\
AnchorPrune adaptation & 0.40 & 0.591 & 0.570 & 0.752 & 0.430 & 0.401 & 0.648 & 0.648 \\
SCOPE pure coverage & 0.40 & 0.587 & 0.630 & 0.804 & 0.370 & 0.401 & 0.623 & 0.623 \\
CoIn reimplementation & 0.40 & 0.595 & 0.632 & 0.822 & 0.368 & 0.401 & 0.689 & 0.689 \\
VisionZip adaptation & 0.40 & 0.580 & 0.650 & 0.810 & 0.350 & 0.401 & 1.000 & 0.575 \\
FastV adaptation & 0.40 & 0.500 & 0.000 & 0.000 & 1.000 & 0.401 & 0.494 & 0.494 \\
\midrule
Protected & 0.50 & 0.667 & 0.308 & 0.642 & 0.692 & 0.500 & 1.000 & 1.000 \\
VisionZip adaptation & 0.50 & 0.542 & 0.834 & 0.918 & 0.166 & 0.500 & 1.000 & 0.661 \\
FastV adaptation & 0.50 & 0.500 & 0.000 & 0.000 & 1.000 & 0.500 & 0.604 & 0.604 \\
\bottomrule
\end{tabular*}
\caption{LLaVA external-method development-set budget curve on TextOCR-Hard. P/N avg.\ is merge-aware source-lineage coverage; Anchor avg.\ uses representative output anchors and is identical for non-merging rows. VisionZip's exhaustive contextual merge makes P/N avg.\ 1.000 without implying separable preservation. SCOPE uses public $\alpha=1$; AnchorPrune passes 12-case exact-index parity; CoIn is reimplemented from the published algorithm. FastV predicts no for every probe and therefore remains at chance accuracy.}
\label{tab:s-llava-official}
\end{table*}

\textbf{Implementation scope.}
We include an external row only when its selector can be executed in our common likelihood-scoring path with a documented parity level. AnchorPrune uses a pinned adaptation of its public LLaVA implementation: only the model interface is changed, while CLIP query priority, vision-CLS attention, two-stage novelty, and native-order selection are retained; 12 deterministic tensor cases reproduce the public selector's indices. CoIn is reimplemented from its published LLaVA-1.5 algorithm and setting. Qwen VisionZip uses a native matched-budget adaptation. Methods whose defining behavior depends on unavailable language-model-internal states are discussed in Related Work but are not replaced by approximate result rows. Table~\ref{tab:s-scope-timing} measures online selection cost, and the artifact records implementation revisions and parity evidence.

\begin{table*}[!tbp]
\centering
\small
\begin{tabular*}{\textwidth}{@{\extracolsep{\fill}}lrrrrrr@{}}
\toprule
Method & Keep & Vision ms & Select/materialize ms & LLM ms & Total ms & Speedup \\
\midrule
Full prefix & 1.000 & 9.9$\pm$0.2 & 0.9$\pm$0.0 & 63.6$\pm$0.1 & 75.4$\pm$0.3 & 1.00$\times$ \\
Protected (40\%) & 0.400 & 10.4$\pm$0.2 & 6.1$\pm$0.1 & 38.0$\pm$0.3 & 55.3$\pm$0.5 & 1.36$\times$ \\
Target (40\%) & 0.401 & 9.9$\pm$0.4 & 5.2$\pm$0.1 & 37.9$\pm$0.3 & 53.9$\pm$0.9 & 1.40$\times$ \\
AnchorPrune (40\%) & 0.401 & 10.0$\pm$0.0 & 13.4$\pm$0.1 & 37.8$\pm$0.0 & 62.0$\pm$0.1 & 1.22$\times$ \\
SCOPE (40\%) & 0.400 & 9.7$\pm$0.2 & 42.2$\pm$1.0 & 37.6$\pm$0.2 & 90.7$\pm$1.3 & 0.83$\times$ \\
CoIn (40\%) & 0.400 & 10.0$\pm$0.2 & 84.6$\pm$1.8 & 37.5$\pm$0.1 & 133.0$\pm$2.0 & 0.57$\times$ \\
\bottomrule
\end{tabular*}
\caption{Repeated exclusive LLaVA single-sample timing on one A800 GPU. Values are mean$\pm$sample standard deviation across three fresh-process repetitions; each repetition uses the same 100 confirmation probes and discards the first five as warm-up. Total additionally includes input preparation, dispatch, and synchronization outside the three named stages, accounting for the approximately 1 ms residual. Full, Target, and AnchorPrune are rotated together; the other rows use companion runs under the same protocol. Target and AnchorPrune remain faster than Full, whereas SCOPE and CoIn cost more than the saved prefill time.}
\label{tab:s-scope-timing}
\end{table*}

\section{Detector-Source Robustness}

Table~\ref{tab:s-box-source} tests whether box-aware variants depend on ground-truth region boxes. The selector receives one of five box sources: ground-truth boxes, simulated detector-like boxes, EasyOCR detections, heavy jitter, or no boxes. The EasyOCR condition supplies every detected text box; it performs no oracle source-box matching, confidence selection, or TextOCR-guided filtering. PosECR and NegSRC are always audited against the original TextOCR boxes, so the table separates detector effects on positive support from retention of negative confusable-source regions.

\begin{table*}[!tbp]
\centering
\small
\begin{tabular*}{\textwidth}{@{\extracolsep{\fill}}lllrrrrr@{}}
\toprule
Model & Box source & Selector & Acc. & hFPR & Keep & PosECR & NegSRC \\
\midrule
LLaVA & GT boxes & Protected (40\%) & 0.661 & 0.298 & 0.400 & 1.000 & 1.000 \\
LLaVA & Simulated detected-like & Protected (40\%) & 0.660 & 0.328 & 0.400 & 0.896 & 0.919 \\
LLaVA & EasyOCR detected & Protected (40\%) & 0.639 & 0.322 & 0.400 & 0.657 & 0.652 \\
LLaVA & Heavy jitter & Protected (40\%) & 0.657 & 0.334 & 0.400 & 0.977 & 0.980 \\
LLaVA & Missing boxes & Protected (40\%) & 0.658 & 0.324 & 0.400 & 0.402 & 0.393 \\
\midrule
InternVL & GT boxes & Soft evidence (50\%) & 0.647 & 0.175 & 0.500 & 0.902 & 0.883 \\
InternVL & Simulated detected-like & Soft evidence (50\%) & 0.646 & 0.187 & 0.500 & 0.902 & 0.874 \\
InternVL & EasyOCR detected & Soft evidence (50\%) & 0.649 & 0.157 & 0.500 & 0.899 & 0.876 \\
InternVL & Heavy jitter & Soft evidence (50\%) & 0.649 & 0.179 & 0.500 & 0.900 & 0.878 \\
InternVL & Missing boxes & Soft evidence (50\%) & 0.642 & 0.179 & 0.500 & 0.887 & 0.854 \\
\bottomrule
\end{tabular*}
\caption{Box-source robustness on TextOCR-Hard. Selection uses the listed box source, while PosECR and NegSRC are audited against original TextOCR boxes with their distinct semantics. EasyOCR supplies all detections without oracle matching or filtering. InternVL rows share one fixed risk-constrained threshold so this table isolates box-source sensitivity rather than calibration. EasyOCR averages 254.4 ms/image, with 56.4\% oracle-box missing rate and 0.155 mean best IoU.}
\label{tab:s-box-source}
\end{table*}

\section{OCRBench, TextVQA, and DocVQA Generalization}

OCRBench tests transfer to a separate OCR-heavy source. On a pruning-compatible 200-probe yes/no diagnostic, Full/moderate-retention accuracy is 0.975/0.920 for Qwen, 0.635/0.640 for LLaVA, and 0.935/0.925 for InternVL; the complete non-native breakdown is included in the artifact. In a complementary 100-question gold-versus-decoy ranking check, Qwen decreases from 0.950 with the full prefix to 0.900 at 30\% retention. Table~\ref{tab:s-ocrbench-generation} retains the more informative native original-question setting with greedy free-form generation; its budget-recovery curve shows that 70\% retention approaches Full more closely than aggressive 30--50\% pruning.

\begin{table*}[!tbp]
\centering
\small
\begin{tabular*}{\textwidth}{@{\extracolsep{\fill}}lrrrrrrrrr@{}}
\toprule
Method & $n$ & Keep & Full exact & Pruned exact & $\Delta$ exact & Full ANLS & Pruned ANLS & $\Delta$ ANLS & Pruned contains \\
\midrule
Qwen Grid (30\%) & 100 & 0.300 & 0.690 & 0.480 & -0.210 & 0.777 & 0.619 & -0.158 & 0.500 \\
Qwen Grid (50\%) & 100 & 0.500 & 0.690 & 0.520 & -0.170 & 0.777 & 0.658 & -0.119 & 0.550 \\
Qwen Grid (70\%) & 100 & 0.700 & 0.690 & 0.660 & -0.030 & 0.777 & 0.736 & -0.041 & 0.680 \\
\bottomrule
\end{tabular*}
\caption{OCRBench original-question greedy generation for Qwen. The selector uses the question text only, never the gold answer string. The result is a boundary check: original-question free-form generation needs a higher retention budget than binary verification or answer ranking.}
\label{tab:s-ocrbench-generation}
\end{table*}

\textbf{Open-QA quality and matched controls.}
Separate 500-sample TextVQA and DocVQA ablation splits established the 30/50/70\% budget range before full validation. Table~\ref{tab:s-openqa-full-validation} scales the frozen question-only selectors to all 5000 TextVQA and 5349 DocVQA validation samples for Qwen and LLaVA. Every pruned-versus-Full interval excludes zero: 70\% recovers substantial quality over aggressive retention, with a steeper curve for LLaVA. Table~\ref{tab:s-openqa-matched-controls} adds matched Random and Grid controls. Target conditioning is not uniformly advantageous in open QA: Random leads on three model--task pairs, and Qwen Target+Grid significantly exceeds Grid only on DocVQA. A question-normalization pilot that removes fixed interrogatives and prompt boilerplate improves DocVQA-500 at 30\% by +0.051 [0.019,0.083], while its other measured intervals include zero; the full Focus comparison is included in the artifact.

\begin{table*}[!tbp]
\centering
\small
\begin{tabular*}{\textwidth}{@{\extracolsep{\fill}}llrrrrrrl@{}}
\toprule
Model & Dataset / metric & $n$ & Keep & Full & Pruned & Paired $\Delta$ & 95\% CI & Win/Loss/Tie \\
\midrule
Qwen3-VL-8B & TextVQA / Acc. & 5000 & 0.300 & 0.8282 & 0.6346 & -0.1936 & $[-0.2057,-0.1816]$ & 183/1269/3548 \\
Qwen3-VL-8B & TextVQA / Acc. & 5000 & 0.700 & 0.8282 & 0.7947 & -0.0335 & $[-0.0403,-0.0270]$ & 122/308/4570 \\
Qwen3-VL-8B & DocVQA / ANLS & 5349 & 0.300 & 0.9395 & 0.5291 & -0.4103 & $[-0.4228,-0.3976]$ & 91/3101/2157 \\
Qwen3-VL-8B & DocVQA / ANLS & 5349 & 0.700 & 0.9395 & 0.8462 & -0.0933 & $[-0.1015,-0.0855]$ & 94/944/4311 \\
\midrule
LLaVA-1.5-7B & TextVQA / Acc. & 5000 & 0.401 & 0.4847 & 0.2766 & -0.2081 & $[-0.2203,-0.1958]$ & 150/1279/3571 \\
LLaVA-1.5-7B & TextVQA / Acc. & 5000 & 0.701 & 0.4847 & 0.3588 & -0.1259 & $[-0.1364,-0.1155]$ & 139/832/4029 \\
LLaVA-1.5-7B & DocVQA / ANLS & 5349 & 0.401 & 0.2157 & 0.1181 & -0.0976 & $[-0.1063,-0.0889]$ & 183/842/4324 \\
LLaVA-1.5-7B & DocVQA / ANLS & 5349 & 0.701 & 0.2157 & 0.1619 & -0.0538 & $[-0.0616,-0.0461]$ & 221/618/4510 \\
\bottomrule
\end{tabular*}
\caption{Full-validation original-question generation with question-only selectors. Qwen uses Target+Grid at 30\% and 70\%; LLaVA uses Target at 40\% and 70\%. Full answers are generated once per model--dataset pair and reused across budgets. Intervals use 20,000 paired bootstrap samples. The fixed selectors and decoding protocol target external validity and compression risk rather than leaderboard-specific tuning.}
\label{tab:s-openqa-full-validation}
\end{table*}

\begin{table*}[!tbp]
\centering
\small
\begin{tabular*}{\textwidth}{@{\extracolsep{\fill}}llrrrrrr@{}}
\toprule
Model & Dataset & Query & Rand. & Grid & Q--R & 95\% CI & Q--G / 95\% CI \\
\midrule
Qwen3-VL-8B & TextVQA & 0.7947 & 0.8102 & 0.7957 & -0.0156 & $[-0.0231,-0.0081]$ & -0.0011 / $[-0.0086,0.0065]$ \\
Qwen3-VL-8B & DocVQA & 0.8462 & 0.8604 & 0.8218 & -0.0142 & $[-0.0239,-0.0046]$ & +0.0244 / $[0.0131,0.0355]$ \\
LLaVA-1.5-7B & TextVQA & 0.3588 & 0.4491 & 0.4382 & -0.0902 & $[-0.1011,-0.0796]$ & -0.0793 / $[-0.0904,-0.0685]$ \\
LLaVA-1.5-7B & DocVQA & 0.1619 & 0.1819 & 0.1893 & -0.0200 & $[-0.0276,-0.0123]$ & -0.0274 / $[-0.0353,-0.0195]$ \\
\bottomrule
\end{tabular*}
\caption{Matched-budget full-validation controls at nominal 70\% retention. Query-guided denotes Target+Grid for Qwen and Target for LLaVA. Actual mean keep is 0.7000 for Qwen and 0.7014 for LLaVA for every compared selector. Full-prefix answers, decoding, sample order, and materialization are shared; intervals use 20,000 paired bootstrap samples. The result limits the transfer claim: target conditioning is effective for the paper's target-verification setting, but is not a generally superior open-QA policy.}
\label{tab:s-openqa-matched-controls}
\end{table*}

\textbf{Multi-region evidence audit.}
Table~\ref{tab:s-manual-multiregion} reports the final human-corrected 96-sample audit, comprising 48 rows drawn from each 500-sample ablation split; we denote these subsets TextVQA-48 and DocVQA-48. External TextVQA boxes and public DocVQA OCR-derived answer/line-context boxes are retained only as scoped source checks in Table~\ref{tab:s-openqa-stress-bbox}; none is supplied to the box-free selector. Table~\ref{tab:s-openqa-detector-in-loop} separately tests detector-provided multi-region boxes in the pruning loop. The final 96-row set averages 2.02 regions per sample and has no unresolved, empty, invalid, or unlabeled rows. In tables that pool both tasks, score is the equally weighted macro-average of TextVQA soft accuracy and DocVQA ANLS. These experiments quantify evidence availability across increasingly open settings; they are not leaderboard-tuned runs or tests of unique causal use.

\begin{table*}[!tbp]
\centering
\small
\begin{tabular*}{\textwidth}{@{\extracolsep{\fill}}llrrrrrrl@{}}
\toprule
Scope & Keep & $n$ & ECR & Worst & All $\geq$0.50 & Pruned score & $\Delta$score & Reading \\
\midrule
All & 30\% & 96 & 0.297 & 0.216 & 0.105 & 0.346 & -0.653 & aggressive pruning loses most complete support \\
All & 50\% & 96 & 0.507 & 0.385 & 0.354 & 0.465 & -0.534 & average coverage masks weak regions \\
All & 70\% & 96 & 0.729 & 0.613 & 0.729 & 0.703 & -0.296 & higher retention improves support and answers \\
DocVQA-48 & 70\% & 48 & 0.758 & 0.594 & 0.708 & 0.672 & -0.328 & labels and layout anchors remain demanding \\
TextVQA-48 & 70\% & 48 & 0.701 & 0.632 & 0.750 & 0.733 & -0.265 & scene-text evidence is better preserved \\
\bottomrule
\end{tabular*}
\caption{Human-corrected multi-region audit of precomputed Qwen Target+Grid masks. ECR measures union coverage; Worst is the minimum per-region ECR; All $\geq0.50$ requires every annotated region to reach 0.50 ECR. Across 288 sample-budget observations, worst-region ECR has Spearman correlation 0.393 with score change, stronger than mean ECR at 0.327. This is evidence availability, not proof that every annotated region was causally used.}
\label{tab:s-manual-multiregion}
\end{table*}

The following slices diagnose specific stressors rather than an exhaustive robustness claim: they do not cover all dense, multilingual, handwritten, or rotated-text conditions.

\begin{table*}[!tbp]
\centering
\small
\begin{tabular*}{\textwidth}{@{\extracolsep{\fill}}p{0.07\textwidth}p{0.21\textwidth}rrrrp{0.27\textwidth}@{}}
\toprule
Audit & Slice or source & $n$ & Keep & Reference & Test & Reading \\
\midrule
Stress & TextVQA numeric answers & 142 & 30/70\% & 0.850 & 0.548 / 0.780 & 30\% is unsafe; 70\% reduces the drop \\
Stress & DocVQA long questions & 168 & 30/70\% & 0.946 & 0.473 / 0.849 & long document questions need higher retention \\
Stress & DocVQA multi-token answers & 224 & 30/70\% & 0.947 & 0.520 / 0.875 & multi-token answers remain a hard slice \\
\midrule
Boxes & TextVQA external GT boxes & 47 & 70\% & -- & ECR 0.677; all 0.851 & single-region evidence is mostly retained \\
Boxes & DocVQA answer-token boxes & 47 & 70\% & -- & ECR 0.687; all 0.511 & answer spans improve with retention \\
Boxes & DocVQA line-context boxes & 47 & 70\% & -- & ECR 0.773; all 0.298 & multi-region context is stricter \\
\midrule
Noise & TextVQA 25\% coordinate jitter & 47 & 70\% & 0.677 & 0.673 & localization jitter has little effect \\
Noise & DocVQA 25\% coordinate jitter & 47 & 70\% & 0.773 & 0.766 & localization jitter has little effect \\
Noise & TextVQA 40\% box dropout & 47 & 70\% & 0.677 & 0.299 & missing boxes sharply reduce adjusted ECR \\
Noise & DocVQA 40\% box dropout & 47 & 70\% & 0.773 & 0.442 & missing boxes are more harmful than jitter \\
\bottomrule
\end{tabular*}
\caption{Open OCR/document-QA stress and evidence-source audits. Stress rows report full-prefix and pruned scores. Box rows use precomputed Qwen masks with external TextVQA boxes or public DocVQA OCR-derived answer/line-context boxes; ``all'' is the fraction whose every region reaches ECR 0.50. Noise rows perturb only the audit boxes. None of these boxes is supplied to the box-free selector, and the results are diagnostics rather than leaderboard or end-to-end noisy-detector claims.}
\label{tab:s-openqa-stress-bbox}
\end{table*}

\begin{table*}[!tbp]
\centering
\small
\begin{tabular*}{\textwidth}{@{\extracolsep{\fill}}lrrrrrrrrr@{}}
\toprule
Scope & $n$ & Full & Base grid & OCR soft & Metadata grid & $\Delta$base & $\Delta$meta & Boxes & Box rows \\
\midrule
All & 96 & 0.999 & 0.703 & 0.771 & 0.746 & +0.068 & +0.024 & 42.250 & 92 \\
DocVQA-48 & 48 & 1.000 & 0.672 & 0.739 & 0.740 & +0.067 & -0.001 & 69.708 & 48 \\
TextVQA-48 & 48 & 0.998 & 0.733 & 0.802 & 0.752 & +0.069 & +0.050 & 14.792 & 44 \\
Rows with boxes & 92 & 0.999 & 0.701 & 0.772 & 0.746 & +0.071 & +0.025 & 44.087 & 92 \\
\bottomrule
\end{tabular*}
\caption{Open-QA EasyOCR detector-in-loop readout on the fixed TextVQA-48/DocVQA-48 audit set. Base grid is the question-only grid selector on the original inputs; OCR soft adds all EasyOCR boxes as a soft selection prior; Metadata grid reruns the box-enriched inputs while ignoring those boxes, controlling for input and metadata differences. The result shows a scoped detector-box benefit over question/grid controls, not full-prefix recovery or online detector-assisted speedup.}
\label{tab:s-openqa-detector-in-loop}
\end{table*}

\textbf{Annotation reliability and ethics.}
Three human annotators followed a written schema requiring minimal visible answer support, separate boxes for contiguous regions, and typed answer/context roles. The primary annotator corrected proposed regions for all rows. Independent annotation covers 32 rows: a second annotator labeled 12 calibration rows and a third labeled a disjoint 20-row extension without access to the primary boxes. Before adjudication, box-count and label-type sets match on 12/32 and 16/32 rows; mean matched-box/union IoU is 0.393/0.389, directional union coverage is 0.621/0.563, and only 3/32 rows pass the strict box criterion. TextVQA union IoU is 0.441 versus 0.337 for DocVQA. Answer-value presence is substantially more stable (30/32 shared; F1 0.968) than context-role boundaries: field-label F1 is 0.667, while other context categories range from 0 to 0.400. The original 12 rows adopt the reviewed secondary boxes; the extension remains a pre-adjudication reliability audit. Non-author student peers were invited to this finite technical annotation task and participated voluntarily without monetary compensation. The task used public benchmark content and collected no personal or sensitive information; it was treated as low risk and not requiring institutional ethics review. Full agreement tables are included in the artifact.

\section{Adaptive Budget Analysis}

Split-safe pilots select policies on development partitions and apply them unchanged to held-out examples. The gate requires both task scores within 0.01 of fixed 70\%, mean keep at most 0.60, and overhead-aware cost below 0.70. No candidate satisfies all three constraints: selector-side fallback lowers hFPR but loses accuracy and coverage, cross-task policies exceed the retention limit, and sequential answer agreement meets quality only by executing multiple prefixes. Thus, these pilots do not yield a unified efficient controller, and fixed 70\% remains the feasible open-QA point under this contract. The complete controller matrix is included in the artifact as a negative boundary analysis, not a proposed contribution.

\section{Efficiency and Spatial Transfer}

The main paper reports aggregate CUDA throughput and memory from actual shortened-prefix execution. Detector-free single-sample TTFT speedups are 1.45/1.37$\times$ for Qwen Target at 20/30\%, 1.24$\times$ for LLaVA Protected (40\%), and 1.54$\times$ for InternVL Soft evidence (50\%); the corresponding batch-prefill gains are 4.32/3.12, 2.34, and 2.24$\times$. For Qwen Target (20\%), measured/estimated total speedup is 1.065$\times$ at 32 generated tokens and 1.018$\times$ at 128 tokens; a separate high-resolution 32-token run gives 1.055$\times$. Decode length dilutes the prefill gain, and online EasyOCR can erase it: adding its 254.4 ms mean latency makes LLaVA Protected and InternVL Soft evidence slower overall at 32 tokens (0.830$\times$ and 0.907$\times$). Table~\ref{tab:s-scope-timing} separately reports three fresh-process selector-cost repetitions over the same 100 probes, with rotated order and five warm-up probes discarded.

The GSR-Bench boundary remains negative. Its spatial-aware diagnostic reserves 35\% of the budget for subject/object center and overlap anchors, 25\% for relation-conditioned context, and fills the remainder by global grid coverage. On the shared 100-probe slice, InternVL Full/Grid/Spatial-aware accuracy is 0.710/0.530/0.550; spatial-aware raises ECR from Grid's 0.746 to 0.880 but leaves hFPR at 0.680. LLaVA Full/Grid/Spatial-aware accuracy is 0.510/0.530/0.520; ECR rises from 0.408 to 0.698 while hFPR changes from 0.900 to 0.940. On the larger 880-probe split, Full/Grid accuracy is 0.633/0.551 for InternVL and 0.535/0.534 for LLaVA. Higher region retention therefore does not reliably restore full-prefix spatial reasoning.

\section{Position-ID Policy Sensitivity}

We isolate logical position handling from token selection. Compact is the ordinary shortened-prefix path used by the corresponding selector run, in which the model assigns consecutive logical positions. Preserve retains the pre-pruning logical IDs while keeping physical cache slots contiguous. This is a prefill-only GAP-style sensitivity check, not a reproduction of GAP's full generation path. We export fixed per-sample masks from the LLaVA development Protected run and the InternVL held-out Soft-evidence run, then evaluate them under Preserve with the same probes and bfloat16 configuration. Kept-index, ECR, and sample-ID mismatches are therefore zero rather than merely controlled in expectation.

Position handling changes answer behavior despite identical evidence availability. On all 1000 LLaVA development probes, Preserve changes 279 raw decisions and changes accuracy/hFPR from the Compact 0.661/0.298 to 0.654/0.362. On InternVL's 536 held-out probes, own-development calibration gives 0.653/0.328 for Compact and 0.647/0.343 for Preserve; test AUROC changes from 0.696 to 0.678. Applying the Compact-derived threshold to both policies yields 101 decision flips and Preserve accuracy/hFPR of 0.647/0.272. These changes are position-policy effects, not selector or ECR gains. We therefore disclose the policy and do not compare rows that silently use different position semantics.

\end{document}